\documentclass{article} %
\usepackage[final]{colm2026_conference}

\usepackage{microtype}
\usepackage{hyperref}
\usepackage{url}
\usepackage{booktabs}

\usepackage{lineno}

\definecolor{darkblue}{rgb}{0, 0, 0.5}
\hypersetup{colorlinks=true, citecolor=darkblue, linkcolor=darkblue, urlcolor=darkblue}

\usepackage{relsize}
\usepackage{amsmath}
\usepackage{booktabs}
\usepackage{amssymb}
\usepackage{multirow}
\usepackage{multicol}
\usepackage{wrapfig}
\usepackage{tcolorbox}
\tcbuselibrary{breakable}
\tcbuselibrary{raster,skins}
\tcbuselibrary{breakable}
\usepackage{tabularx}
\usepackage{makecell}
\usepackage{listings}
\usepackage{subcaption}
\usepackage{paralist}
\usepackage{booktabs}
\usepackage{xcolor}
\usepackage{colortbl}
\usepackage{tablefootnote}
\usepackage{comment}
\usepackage{enumitem}
\usepackage{paralist}
\usepackage{comment}
\usetikzlibrary{positioning, shapes.geometric, arrows.meta}

\definecolor{deeppurple}{RGB}{28, 109, 172}
\definecolor{cornell_red}{RGB}{153,51,102} 
\definecolor{interpretcolor}{RGB}{191,87,0}

\definecolor{prompt_colour}{RGB}{76,87,133}
\definecolor{dutch_white}{RGB}{234, 222, 189}
\definecolor{prompt_colour2}{RGB}{159, 167, 200}
\definecolor{cool_gray}{RGB}{150,150,150}

\definecolor{clarif_colour}{RGB}{165,88,119}
\definecolor{user_sim_colour}{RGB}{52,118,105}
\definecolor{answer_gen_colour}{RGB}{29,99,129}

\makeatletter
\newcommand{\tightparagraph}{\@startsection{paragraph}{4}{\z@}{0.01ex}{-1em}{\normalsize\bf}}
 \makeatother

\title{Reasoning about Intent for Ambiguous Requests}

\author{Irina Saparina  \& Mirella Lapata \\  
  School of Informatics, \\
  University of Edinburgh  \\
   Edinburgh, UK\\
    \texttt{i.saparina@sms.ed.ac.uk}, \  \texttt{mlap@inf.ed.ac.uk}}

\begin{document}

\ifcolmsubmission
\linenumbers
\fi

\maketitle

\begin{abstract}

Large language models often respond to ambiguous requests by \emph{implicitly} committing to one interpretation, frustrating users and creating safety risks when that interpretation is wrong. 
We propose generating a single structured response that enumerates the different ways an ambiguous request can be interpreted, each coupled with a corresponding answer.
Our models are trained with reinforcement learning using a dual reward objective: recall on ambiguous inputs to maximise coverage of valid interpretations, and precision on unambiguous ones to suppress spurious alternatives.
Training requires only multiple valid answers per input as supervision, no clarification questions or explicit interpretations are needed. 
Experiments on conversational question answering and semantic parsing demonstrate that our method achieves higher coverage of valid answers than baseline approaches. 
Human evaluation confirms that predicted interpretations are meaningful and explain their corresponding answers.  
Our approach promotes transparency with \emph{explicit} interpretations, 
avoids additional interaction rounds by producing all interpretations in a
single generation,
and supports downstream applications through its structured output format.

\end{abstract}

\section{Introduction}
Ambiguity poses a fundamental challenge for large language models
(LLMs). Users routinely omit details either for the sake of brevity or
unintentionally \citep{zipf.g.1949, piantadosi2012communicative},
creating requests with multiple plausible interpretations (see the
examples in Table~\ref{tab:types_of_response}).  LLMs often
respond with a single answer, implicitly committing to one
interpretation without acknowledging alternatives.
LLMs can
strategically exploit ambiguities, e.g.,~by deliberately
misinterpreting scalar quantifiers or bracketing in logical statements
to advance their own goals over user intentions
\citep{choi2025language}. Ambiguity also presents a challenge for
reasoning models, which can expend thousands of tokens on incorrect
interpretations, simultaneously wasting computational resources and
reinforcing the original error.

Prior work has explored different strategies for handling ambiguous
inputs, each with distinct trade-offs
\citep{tanjim2025disambiguation}. 
Clarification questions resolve ambiguity,
but require additional interaction rounds 
(see Table~\ref{tab:types_of_response}). An alternative is to generate a 
longer answer that covers multiple interpretations in a single
response \citep{stelmakh-etal-2022-asqa}. 
Outputs in this case tend to be verbose, lacking explicit interpretations (see the examples in Table~\ref{tab:types_of_response}). As a result, the answers are difficult to parse for downstream applications and potentially overwhelming for users. 
Moreover, for certain ambiguity types (e.g.,
attachment ambiguity; see Table~\ref{tab:types_of_response}, right), interpretations may be mutually contradictory. In
domains like SQL generation, there may be no natural way to express a
unified response (e.g., for interpretations requiring different table structures).

\begin{table}[t]
\small
\begin{tcbraster}[
  raster columns=2,
  raster column skip=6mm,
  raster equal height=rows,  %
  raster valign=top
]
\begin{tcolorbox}[
  enhanced,
  colback=white,
  colframe=black!75,
  boxrule=1pt,
  arc=2mm,
  left=1mm,right=1mm,top=1mm,bottom=1mm
]
\centerline{\textbf{Conversational Question Answering}}
\vspace{1mm}\hrule\vspace{1mm}
{\fontsize{8}{9.5}\selectfont
\textcolor{deeppurple}{ The first states to hold power in Silesia were \textcolor{cornell_red}{Greater Moravia} in the 9th c. and \textcolor{cornell_red}{Bohemia}  in the 10th c. Later, Silesia became a
part of \textcolor{cornell_red}{the early Polish state} and in the 14th c., \textcolor{cornell_red}{the Holy Roman Empire}...}
}
\fontsize{8}{9.5}\selectfont
\par\vspace{1mm}
\begin{tabular}{@{}l@{\hspace{10mm}}l@{}}
\textcolor{deeppurple}{-- When did Bohemia rule there?} & \textcolor{deeppurple}{-- 10th c.} \\
\textcolor{cornell_red}{-- What other state ruled it?} & \\
\end{tabular}
\vspace{1mm}\hrule\vspace{1mm}

\centerline{\textit{\small Clarification Question}}
\vspace{1mm}\hrule\vspace{1mm}
\fontsize{8}{9.5}\selectfont
\begin{tabular}{@{}l@{}l@{}}
-- Do you mean what state ruled it in the 9th c.? & \\
\textcolor{cool_gray}{-- No, I mean  what state ruled it in the 14th c.} & \\
\textcolor{cool_gray}{-- The Holy Roman Empire} & \\
\end{tabular}
\vspace{1mm}\hrule\vspace{1mm}
\centerline{\textit{\small Long-form Answer}}
\vspace{1mm}\hrule\vspace{1mm}
{\fontsize{8}{9.5}\selectfont
Greater Moravia ruled in the 9th c., and later the Polish state, and the Holy Roman Empire.}
\vspace{1mm}\hrule\vspace{1mm}
\centerline{\textit{\small Interpretation-Answer Pairs (Ours)}}
\vspace{1mm}\hrule\vspace{1mm}
\begin{enumerate}[leftmargin=*,nosep]
\fontsize{8}{9.5}\selectfont
\item \textcolor{interpretcolor}{Other state refers to the predecessor of Bohemia}\\
\hspace{.4cm}\textcolor{deeppurple}{Greater Moravia}
\item \textcolor{interpretcolor}{Other state refers to the successor of Bohemia}\\
\hspace{.4cm}\textcolor{deeppurple}{The early Polish state}
\item \textcolor{interpretcolor}{Other state refers to the 14th-century ruler}\\
\hspace{.4cm}\textcolor{deeppurple}{The Holy Roman Empire}
\end{enumerate}
\end{tcolorbox}
\begin{tcolorbox}[
  enhanced,
  colback=white,
  colframe=black!75,
  boxrule=1pt,
  arc=2mm,
  left=1mm,right=1mm,top=1mm,bottom=1mm
]
\centerline{\textbf{Text-to-SQL Parsing}}
\vspace{1mm}\hrule\vspace{1.5mm}
\begin{lstlisting}[escapeinside=@@,aboveskip=0pt,belowskip=0pt,
basicstyle=\ttfamily\fontsize{6.7}{8}\selectfont\color{deeppurple},xleftmargin=0pt,framexleftmargin=0pt,breaklines=true]
CREATE TABLE Films(
    id INTEGER, title TEXT, @\textcolor{cornell_red}{budget}@ REAL);
\end{lstlisting}
\vspace{1mm}
\begin{lstlisting}[escapeinside=@@,aboveskip=0pt,belowskip=0pt,
basicstyle=\ttfamily\fontsize{6.7}{8}\selectfont\color{deeppurple},xleftmargin=0pt,framexleftmargin=0pt,breaklines=true]
CREATE TABLE Genres(
    id INTEGER, @\textcolor{cornell_red}{genre}@ TEXT);
\end{lstlisting}
\vspace{1mm}
\begin{lstlisting}[escapeinside=@@,aboveskip=0pt,belowskip=0pt,
basicstyle=\ttfamily\fontsize{6.7}{8}\selectfont\color{deeppurple},xleftmargin=0pt,framexleftmargin=0pt,breaklines=true]
CREATE TABLE Film_Genre ...
\end{lstlisting}\vspace{1mm} 
\textcolor{cornell_red}{\fontsize{8}{9.5}\selectfont Show horror films and thrillers with a 20M budget}
\vspace{1mm}\hrule\vspace{1mm}
\centerline{\textit{\small Clarification Question}}
\vspace{1mm}\hrule\vspace{1mm}
\begin{itemize}[nosep,label=--,leftmargin=*]
\item {\fontsize{8}{9.5}\selectfont Does the budget apply to both genres?}
\item[\textcolor{cool_gray}{--}] {\fontsize{8}{9.5}\selectfont \textcolor{cool_gray}{Only to thrillers.}}
\item[\textcolor{cool_gray}{--}] \textcolor{cool_gray}{\fontsize{6.7}{8}\selectfont{\texttt{SELECT title ... WHERE genre = `Horror' OR}}}\\[-0.5mm]
\textcolor{cool_gray}{\fontsize{6.7}{8}\selectfont{\texttt{genre = `Thriller' AND budget = 20;}}}
\end{itemize}
\vspace{1mm}\hrule\vspace{1.2mm}
\centerline{\textit{\small Long-form Answer}}
\vspace{1mm}\hrule\vspace{1.5mm}
{\fontsize{8}{9.5}\selectfont N/A}
\vspace{1.5mm}\hrule\vspace{1mm}
\centerline{\textit{\small Interpretation-Answer Pairs (Ours)}}
\vspace{1mm}\hrule\vspace{1.5mm}
\begin{enumerate}[leftmargin=*,nosep]
\item {\textcolor{interpretcolor}{\fontsize{8}{9.5}\selectfont Budget applies to both genres}}\\
\hspace{.4cm}\textcolor{deeppurple}{\fontsize{6.7}{8}\selectfont\texttt{SELECT title  ... WHERE (genre = `Horror' OR}}\\[-0.5mm]
\textcolor{deeppurple}{\fontsize{6.7}{8}\selectfont\texttt{genre = `Thriller') AND budget = 20}}
\item {\textcolor{interpretcolor}{\fontsize{8}{9.5}\selectfont Budget applies only to thrillers}}\\
\hspace{.4cm}\textcolor{deeppurple}{\fontsize{6.7}{8}\selectfont\texttt{SELECT title ... WHERE genre = `Horror' OR}}\\[-0.5mm]
\textcolor{deeppurple}{\fontsize{6.7}{8}\selectfont\texttt{genre = `Thriller' AND budget = 20;}}
\end{enumerate}
\end{tcolorbox}
\end{tcbraster}
\caption{
Approaches to handling ambiguity. Examples (simplified for the sake of brevity) are from the Abg-CoQA \citep{guo2021AbgCoQA} and Ambrosia \citep{ambrosia} test sets, illustrating time-dependent ambiguity (left) and attachment ambiguity (right).
Inputs and ground-truth answers are shown in \textcolor{deeppurple}{blue}, ambiguous questions and elements that create ambiguity are highlighted in \textcolor{cornell_red}{red}. 
Interaction rounds following clarification questions are shown in \textcolor{cool_gray}{gray}. Our approach pairs each answer with an explicit interpretation (shown in \textcolor{interpretcolor}{orange}) of the ambiguous question. 
} 
\label{tab:types_of_response}
\vspace{-3mm}
\end{table}

In this work, we propose a structured alternative, namely generating
multiple plausible answers, each paired with an explicit
interpretation  (Table~\ref{tab:types_of_response}, bottom). This
format requires  a single generation step (rather than a resolving
dialogue or a multi-stage pipeline) while producing outputs
that are easy to parse and use for downstream applications, e.g.,~in
agentic systems where branching across interpretations can be
beneficial.
Moreover, it extends beyond textual answers to other forms of output, e.g.,~code, while allowing users to
quickly identify their intended answer.
Recognizing that ambiguity relates to the broader problem of pluralistic
alignment (i.e., generating responses that reflect diverse
perspectives and interpretations; \citealt{sorensen2024}), our approach
promotes \emph{overton pluralism} by predicting multiple
interpretations within one answer. Methods embracing
\emph{distributional pluralism} rely on parallel samples and often
struggle to produce substantially different outputs
\citep{zhang2025noveltybench}. Instead, we generate
interpretations sequentially within one response (where subsequent
interpretations are conditioned on previous ones), without changes to sampling   or confidence calibration. %

Teaching models to handle ambiguity is challenging due to the lack of
suitable supervision. While ambiguity is pervasive in natural language,
humans can communicate without  clarifying
their intent. %
The most readily available supervision signal comes from
annotator disagreement, which manifests as multiple answers for a
single instance. Other forms of annotation, such as clarifications or explicit interpretations, are expensive and difficult to
obtain at scale. Building on the success of reinforcement learning
under sparse supervision \citep{deepseek-math}, we frame ambiguity
resolution as reasoning about \textit{what} to answer rather than
\textit{how}: we encourage the model to reason about user intent, entertaining different interpretations of the input before committing
to an answer. 
We train %
a reasoning model through reinforcement
learning with specialized reward functions that only require answer
annotations as supervision. For ambiguous inputs, we optimize for
\emph{recall} to maximize the coverage of interpretations. For
unambiguous inputs, we optimize for \emph{precision} to avoid
generating spurious alternatives. This dual objective teaches the
model to enumerate interpretations when necessary \emph{and} to
recognize when a question has a single intent.

We evaluate our approach on both verifiable and non-verifiable tasks, including text-to-SQL generation and conversational question answering. %
Results show that our model effectively handles  unambiguous and ambiguous cases, generalizing to out-of-distribution data while generating diverse, plausible answers with high coverage and meaningful interpretations. By teaching models to reason about user intent before answering, we move toward more robust, user-aligned LLMs that embrace rather than obscure ambiguity.

\section{Related Work}

\paragraph{Ambiguity in Question Answering}
A growing body of work recognises that many questions
admit multiple valid answers, each corresponding to a distinct
interpretation (see \citet{tanjim2025disambiguation} for an overview).
AmbiQA \citep{min-etal-2020-ambigqa}
was one of the first large-scale benchmarks to explicitly annotate
multiple answers for individual questions, expanding on
NaturalQuestions \citep{kwiatkowski-etal-2019-natural}. The
open-domain QA setting has prompted numerous retrieval-augmented
generation techniques (e.g., \citealp{min-etal-2021-joint}).
ASQA \citep{stelmakh-etal-2022-asqa} extends AmbiQA by requiring
long-form answers that cover multiple interpretations in a single
response; while intuitive, this format lacks explicit
structure, complicating both evaluation and downstream use.
\citet{amplayo-etal-2023-query} infer multiple interpretations
and then generate long-form answers, though they rely on prompting only. 
\citet{guo2021AbgCoQA} use
clarification questions as a disambiguation strategy in conversational
QA, and more recently \citet{Chen2024LearningTC} introduce a
preference optimization method that generates clarification questions
for ambiguous instances and direct answers for unambiguous ones. 
\citet{zhang2025modeling} train models to ask clarifying questions by modelling
future conversation turns, while concurrent work learns steerable clarification
policies via collaborative self-play \citep{berant2025learning}.
In contrast, our approach produces structured
interpretation-answer pairs and handles both cases in a \emph{single}
turn, without requiring additional interaction rounds.

\paragraph{Ambiguity in Semantic Parsing}
Ambiguity also presents challenges for semantic parsing
across a range of settings
\citep{rasmussen-schuler-2020-corpus, li-etal-2023-python,
mu2023clarifygpt, eskin-2024-zero}. Standard datasets
often provide only a single interpretation when multiple queries are
equally plausible \citep{pourreza-rafiei-2023-evaluating,
Floratou2024}, motivating benchmarks that include multiple SQL
queries per input \citep{bhaskar-etal-2023-benchmarking, ambrosia} or
interactive scenarios with clarification questions
\citep{dong-etal-2025-practiq, huo2025bird}.
Proposed solutions include counterfactual examples
\citep{wang-etal-2023-know}, specialized decoding strategies
\citep{bhaskar-etal-2023-benchmarking}, generating interpretations and then answers in two stages
\citep{saparina-lapata-2025-disambiguate}, and interactive clarification
\citep{ding2025ambisql}. Unlike multi-stage approaches, our method
generates all plausible interpretations and corresponding SQL queries
in one response.

\paragraph{Diversity in Model Output}
Our work also relates to efforts to improve output
diversity in LLMs, which have been shown to exhibit narrow behaviours
and strong biases even in tasks where many valid responses exist
\citep{ambrosia, west2025base, zhang2025noveltybench,
troshin2025asking}. Post-training approaches have been proposed to shift model outputs from peaked to more diffuse distributions \citep{zhang2024forcing, sorensen2025spectrumtuningposttrainingdistributional},  including concurrent work
that trains models to generate diverse answers for questions admitting multiple
correct ones \citep{puri2026reaching}. \citet{zhang2025verbalized} improve diversity by
enumerating possible answers with explicit probability estimates via
prompting. 
Our approach pursues diversity through a different lens: it targets semantic ambiguity, where interpretations lead to fundamentally different answers, and pairs each answer with an explicit interpretation rather than a probability estimate, making the source of diversity explicit and actionable.

\section{Reasoning about User Intent}

\subsection{Problem Formulation}
 A question is ambiguous if it admits multiple distinct
 interpretations that lead to different answers.  Given some input
 consisting of optional context~$c$ (e.g., a passage in 
 conversational QA or a database schema in text-to-SQL generation; 
see Table~\ref{tab:types_of_response}) and a question~$q$, the
 model must produce a set of interpretation-answer pairs $\{(i_1,
 a_1), \dots (i_m, a_m)\}$. Each interpretation~$i_k$ explicitly
 disambiguates the question by specifying what is being asked,  %
 while~$a_k$ is the corresponding answer for
 interpretation~$i_k$.  
 For example, ``Other state refers to the predecessor of Bohemia'' 
in Table~\ref{tab:types_of_response} is one possible interpretation of 
the ambiguous question ``What other state ruled it?'' which corresponds 
to the answer ``Greater Moravia''.

We expect a maximum of five interpretation-answer pairs, which we find
represents the majority of ambiguity cases in existing datasets (Section~\ref{sec:metrics}). 
In theory, questions could have more interpretations, but at
that scale  other approaches such as clarification
questions may be more appropriate than  enumeration. %
In practice, the questions are often grounded in specific contexts, which naturally constrain the space of plausible interpretations. 
For unambiguous cases,  we expect a single interpretation-answer pair,
with the  interpretation being identical to the original question or a
paraphrase that makes the intent even more explicit.

\subsection{Model Training}
Our model is trained on  examples  annotated with gold answers
$\mathcal{A}=\{\hat{a_1}, \dots \hat{a_n}\}$, where $n=1$ for
unambiguous and $n>1$ for ambiguous questions. These annotations are
 easier to obtain compared to interpretations or clarification
questions; for example, they emerge naturally when multiple annotators
disagree on the answers they provide. 

We instruct the model to first analyze the input for potential
ambiguities, verbalizing its reasoning as a chain of thought, and then generate structured output consisting of
interpretation-answer pairs. We design a task-agnostic prompt template
that can be easily adapted to different tasks by only modifying the
context format and task-specific instructions (e.g., answer style
constraints). The prompt is given in Appendix~\ref{appx:prompts}.
We  generate interpretation-answer pairs sequentially.
Thanks to the autoregressive nature of language models, each subsequent
interpretation is conditioned on previously generated
interpretations and answers. 
This encourages diversity across interpretations  \citep{zhang2025noveltybench, troshin2025asking, zhang2025verbalized} as well as semantic alignment between
interpretations and their corresponding answers. We confirm the diversity effect in
Appendix~\ref{appx:sampling}, where we compare against parallel sampling.

We train our model using DAPO (Decoupled Clip and Dynamic Sampling
Policy Optimization \citealt{yu2025dapo}), a reinforcement learning
algorithm based on GRPO (Group Relative Policy Optimization;
\citealt{deepseek-math}). %
DAPO
extends GRPO with %
decoupled clipping parameters that enhance exploration and
dynamic sampling that filters out prompts with uniform outcomes to
maintain effective gradients (see Appendix~\ref{appx:grpo_dapo} for
the formulas).

We design \textbf{reward functions} to optimize different objectives
for ambiguous versus unambiguous inputs.  %
We extract the set of predicted answers~$\mathcal{P}$ from the model's
output and compare it against the gold answer set~$\mathcal{A}$, where
all answers are distinct by construction.  For each predicted
answer~$p\in\mathcal{P}$ and gold answer~$\hat{a}\in\mathcal{A}$ we
compute a \textbf{similarity} score $sim(p, \hat{a})$ where the
specific type of similarity function is task dependent.

For tasks like text-to-SQL whose output can be objectively verified,
$sim(p, \hat{a}) = 1$ if execution results are the same, and $sim(p,
\hat{a}) = 0$ otherwise. %
For tasks like question answering whose outputs are harder to verify,
we use soft similarity metrics. In experiments, we adopt F1 score
based on word overlap, however, any other word similarity function
could be used. Let~$S_{ij} =sim(p_i, \hat{a}_j)$ denote a matching
matrix, and $\mathcal{M}$~the set of one-to-one matches between predicted and gold answers. 

We compute optimal assignments between
$\mathcal{P}$ and $\mathcal{A}$ as a maximum weighted bipartite matching problem using the Hungarian algorithm~\footnote{  We use the Hungarian algorithm as a general-purpose solution. With a maximum of five interpretations,  brute-force  optimization 
 is also tractable. 
}:
$$M^* =
\text{argmax}_{M\in\mathcal{M}} \sum_{(i, j) \in M} S_{ij}.$$ %
{For verifiable tasks with binary similarity (e.g., text-to-SQL), $M^*$ reduces to exact match.}

For ambiguous questions  ($|\mathcal{A}|>1$),  our reward function is
\textbf{recall}:
 $$R_{\text{recall}}=\dfrac{1}{|\mathcal{A}|}\sum_{(p_i, \hat{a}_j)\in
  M^*}sim(p_i, \hat{a_j})$$
which measures how  many distinct valid interpretations are covered by
the model, rewarding more complete responses. 

For unambiguous questions ($|\mathcal{A}|=1$), our primary objective
is \textbf{precision} which penalizes generating spurious interpretations: 
 $$ R_{\text{precision}}=\dfrac{1}{|\mathcal{P}|}\sum_{(p_i, \hat{a}_j)\in M^*}sim(p_i, \hat{a_j})$$

\section{Experimental Setup}
\subsection{Datasets and Evaluation Metrics}\label{sec:metrics}

The only requirement for training our model is a dataset with multiple valid answers per input. %
We evaluate on three datasets representing different tasks and evaluation scenarios. 
All datasets include contextual information (e.g., passages or databases), making ambiguity resolution challenging and realistic as models must validate their interpretations against the provided context.
Appendix~\ref{appx:types_of_ambiguity} describes the types of ambiguity present in the datasets.

\textbf{Abg-CoQA} \citep{guo2021AbgCoQA} is a conversational
question answering dataset with non-verifiable outputs. The input consists of a passage, one or two previous conversation turns, and a question that can be ambiguous, leading to different answers. 
The dataset provides clarification questions but does not include explicit
interpretations. 
It covers different types of ambiguity, the
majority being coreference, and contains 741 ambiguous and 5,727
unambiguous training examples and 123 ambiguous and 932 unambiguous
test examples. 
Ambiguous questions have 2.21 answers on average, with a maximum of 4.

\citet{guo2021AbgCoQA} measure similarity between predicted and gold
answers using word overlap (macro-averaged F1). To account for
different wordings within the same interpretation, each prediction is
compared against three gold answers, and the maximum score is taken;
these scores are then averaged across gold answers.
This metric favors coverage over precision because
Abg-CoQA treats disambiguation and answer generation as separate
tasks. 
For a more comprehensive evaluation, we also use an LLM-based judge (Qwen3 4B Instruct; \citealp{yang2025qwen3technicalreport})
to assess recall (the proportion of gold answers covered by the
predictions), precision (the proportion of predictions that match a
gold answer), and full coverage (the percentage of instances where
all gold answers are recovered). For unambiguous questions, there is
only one gold answer, so recall and full coverage are equivalent and
we report recall in these cases. Details on our prompt and judge model selection are
in Appendix~\ref{appx:judge}.

\textbf{Ambrosia} \citep{ambrosia} is a text-to-SQL dataset, with verifiable outputs,  where
some user questions are ambiguous and can lead to different SQL
queries and execution results.  
The dataset showcases multiple ambiguities including scope and attachment ambiguities and comes with gold interpretations for ambiguous questions. %
We compare query execution results and report recall, precision, and full coverage. We use the data split from
\citet{saparina-lapata-2025-disambiguate},  containing 1,026 ambiguous and 2,380 unambiguous training examples and 131 ambiguous and 309 unambiguous test examples.  Ambiguous questions have 2.3 answers on average, with a maximum of 3.

\textbf{AmbiQT} \citep{bhaskar-etal-2023-benchmarking} is a text-to-SQL benchmark with ambiguous questions that we use to evaluate out-of-domain generalization, as models trained on Ambrosia are evaluated on AmbiQT
without fine-tuning. Unlike Ambrosia's synthetic databases, AmbiQT features real-world databases with more complex schemas. Nevertheless, generalization is possible because both datasets share vague ambiguity types. We use AmbiQT as a held-out test set: models trained on Ambrosia are evaluated on AmbiQT without fine-tuning. Following \citet{saparina-lapata-2025-disambiguate}, we use the same preprocessed splits (1,954 examples) and report execution match recall, precision, single and full coverage.

\subsection{Training Details}
\label{sec:training-details}

We train Qwen3 4B Instruct \citep{yang2025qwen3technicalreport} as our
base model which offers a good balance between performance and
efficiency; it is small enough for fast training and inference while
maintaining competitive performance. We do not use reasoning models
(e.g., Qwen3 4B Thinking), as preliminary experiments showed that they  do not outperform instruction-tuned models on
ambiguity resolution, while incurring
much higher cost and latency.

\paragraph{Data Balancing} Datasets with ambiguous inputs are often
 imbalanced, with many more unambiguous examples than ambiguous ones.
  However, ambiguous examples
 are more challenging and require stronger learning signals,
 especially with methods based on GRPO that have difficulty-level bias
 \citep{liu2025understanding}. We therefore rebalance the training
 data to maintain an ambiguous-to-unambiguous ratio of 3:1, by 
 oversampling ambiguous examples.
 Ambiguous examples teach the model when to generate multiple
 interpretations, while unambiguous ones serve as regularization,
 preventing it from over-generating interpretations and
 maintaining high precision on straightforward inputs. During
 training, we sample 4 examples per batch, which combined with
 the 3:1 ratio, allocates more compute to harder ambiguous cases while
 still maintaining  signal from unambiguous examples. %

\paragraph{Reward Functions} For Abg-CoQA, we compute rewards using  word-overlap F1 as
the similarity metric. While less reliable than LLM-as-a-judge for
final evaluation, we found F1 provides effective training signal
during reinforcement learning and is much faster to compute. For
Ambrosia, we use execution match (comparing predicted and gold SQL
query results) which provides an objective, verifiable reward signal.
We train using DAPO \citep{yu2025dapo} for 5 epochs with a learning
rate of 5e-7 and linear learning rate scheduling with 3\% warmup. We
generate 16 completions per prompt during training. Additional
hyperparameters are in Appendix~\ref{appx:training_details}.

\subsection{Comparison Methods}
We compare against approaches that isolate key aspects of ambiguity handling, namely prompting-based approaches, supervised fine-tuning, and  state-of-the-art methods relying on interaction or multi-stage generation (implementation details are  in Appendix~\ref{appx:comparison_details}):

\paragraph{IntentPrompt} 
We provide models with explicit instructions to reason about user intent, detect ambiguity and output  multiple interpretation-answer pairs, using the same prompt template as our training approach (see Appendix~\ref{appx:prompts}). This baseline isolates whether ambiguity handling can be achieved purely through prompting, without additional training.
  We evaluate across two  model scales: Qwen3 4B and 235B (22B active parameters), with both standard instruction-tuned (Instruct) and reasoning (Thinking) variants.

\paragraph{SFT}
We fine-tune models  to directly predict multiple %
interpretation-answer pairs. %
On Abg-CoQA we predict answers only  as the dataset lacks gold interpretations.
We use IntentPrompt but exclude reasoning instructions due to the absence of gold chain-of-thought annotations. For fine-tuning, we use Qwen3 4B Instruct, to match our model's scale.

\paragraph{State-of-the-Art} We compare against two state-of-the-art methods.  \textit{ACT} \citep{Chen2024LearningTC} is a  preference optimization
approach that trains models to ask clarification questions for ambiguous
requests, requiring multiple stages to resolve ambiguity.
We compare on Abg-CoQA against their best system, which uses Zephyr 7B~$\beta$ \citep{tunstall2023zephyrdirectdistillationlm} as the policy model and Gemini Ultra~1.0 \citep{geminiteam2025geminifamilyhighlycapable} for
user simulation. \textit{DisambigParse} \citep{saparina-lapata-2025-disambiguate} is a two-stage approach for handling ambiguity in semantic parsing that first generates interpretations (with Llama3.1 8B Instruct; \citealt{Dubey2024TheL3}) and then uses them to guide SQL generation (with Qwen2.5 Coder 32B Instruct; \citealt{hui2024qwen25codertechnicalreport}).

\section{Results}

\subsection{Comparison with Prompting and Fine-tuning}

\paragraph{Conversational Question Answering}
Table~\ref{tab:Abg-CoQA_results} summarizes our results on 
Abg-CoQA using word overlap F1 and
LLM-as-a-Judge  as evaluation metrics.
We observe that larger models consistently
outperform smaller ones, which suggests that our approach would
benefit from stronger backbone models. 
Interestingly, IntentPrompt baselines achieve higher recall than precision, indicating that they over-detect ambiguity. %
For instance, the 235B
Thinking model achieves {66.2\% recall but only 49.7\% precision} on
ambiguous questions; it generates multiple
answers, some of which are spurious.

\begin{table}[t]
\centering
\small
\begin{tabular}{@{}l@{\;\;\;}l@{\;\;\;}c@{\;\;}c@{\;\;}c@{\;\;}c@{\;\;\;}c@{\;\;}c@{\;\;}c@{}}
\toprule
 & & \multicolumn{4}{c}{\textbf{Ambiguous}} & \multicolumn{3}{c}{\textbf{Unambiguous}} \\ 
\cmidrule(lr){3-6} \cmidrule(lr){7-9}
\cmidrule(lr){4-6}
 & & \textbf{Overlap} & \textbf{Judge} & \textbf{Judge} & \textbf{Full} & \textbf{Overlap} & \textbf{Judge} & \textbf{Judge} \\ 
\textbf{Method} & \textbf{Model}  & \textbf{F1} & \textbf{Recall} & \textbf{Precision} & \textbf{Coverage} & \textbf{F1} & \textbf{Recall} & \textbf{Precision} \\ 
\midrule
\multirow{4}{*}{{IntentPrompt}} 
  & 4B Instruct & 44.4 & 62.3  & 42.7 &  45.5 & 59.2 & 77.7 & 40.6  \\
  & 4B Thinking & 18.2 & 46.2 & 45.3 & 31.7 & 32.5 & 60.6 & 44.6 \\
  & 235B Instruct & 55.2 & 63.1 & 47.0 & 45.5 & 68.9 & 80.8 & 42.1 \\
  & 235B Thinking & 47.8 & 66.2 & 49.7 & 50.4 & 59.6 & 82.1 & 47.2 \\
SFT\textsuperscript{*}
  & 4B Instruct & 40.1 &  48.1 & \textbf{70.3} & 20.3  & 72.2 & 77.6 &	\textbf{70.6} \\
\multirow{1}{*}{IntentRL} 

  & 4B Instruct & \textbf{72.9} & \textbf{78.1} & 53.5 & \textbf{61.0}  &  \textbf{84.4} & \textbf{90.9}	 &  40.7 \\
\bottomrule
\end{tabular}
\vspace{-5pt}
\caption{Evaluation on \textbf{Abg-CoQA} using Word-Overlap F1 and LLM-as-a-Judge metrics (Recall, Precision, Full Coverage)  for ambiguous and unambiguous questions. SFT\textsuperscript{*} outputs answers only as gold interpretations are unavailable. Best results per metric are in \textbf{bold}. %
}
\label{tab:Abg-CoQA_results}
\end{table}

Supervised fine-tuning (SFT\textsuperscript{*}, answers only since Abg-CoQA lacks gold interpretations) reveals a critical limitation: despite
our efforts to rebalance the training data and make unambiguous
examples less frequent than ambiguous ones, the model overfits to
unambiguous cases. This results in high precision (70.3\% for
ambiguous, 70.6\% for unambiguous) at the cost of recall (48.1\%) and
full coverage (20.3\%) for ambiguous questions. The
model learns to commit to a single answer rather than enumerate multiple options. 

Our RL-based model (IntentRL) achieves the best
results for ambiguous questions: 72.9\% word-overlap F1, 78.1\%
recall, and 61.0\% full coverage. For unambiguous questions, we
achieve 84.4\% word-overlap F1 and 90.9\% recall, demonstrating that
the model successfully  recognizes when questions have a
single intent. Lower precision  (53.5\% for ambiguous,
40.7\% for unambiguous) reflects a bias toward generating additional
interpretations as the model sometimes treats unambiguous questions as
ambiguous and produces alternatives.  We achieve the best
precision on ambiguous questions among all interpretation-answer
approaches (53.5\% vs.\ 42.7--49.7\% for IntentPrompt), showing that our method
improves precision even though we do not explicitly optimize for it in
the reward function, and the base models initially exhibit even lower
precision than recall. High recall implies that when the
model generates multiple interpretations, the correct ones are often included. In safety-critical applications, over-generating
plausible interpretations is preferable to silently committing to 
one, potentially incorrect interpretation.

\begin{table}[t]
\centering
\small
\begin{tabular}{llccccc}
\toprule
 & & \multicolumn{3}{c}{\textbf{Ambiguous}} & \multicolumn{2}{c}{\textbf{Unambiguous}} \\ 
\cmidrule(lr){3-5} \cmidrule(lr){6-7}
 \textbf{Method}& \textbf{Model} & \textbf{Recall} & \textbf{Precision} & \textbf{Full Coverage} & \textbf{Recall} & \textbf{Precision} \\ 
\midrule
\multirow{4}{*}{{IntentPrompt}} 
  & 4B Instruct & 22.5 & 30.8 & 3.8 & 22.0 & 18.8 \\
  & 4B Thinking & 16.9 & 23.2 & 3.1 & 25.2 & 19.7 \\
  & 235B Instruct & 46.2 & 48.9 & 18.3 & 74.1 & 57.8  \\
  & 235B Thinking & 37.3 & 37.8 & 12.2 & 53.7 & 35.9 \\
\multirow{1}{*}{SFT} 
  & 4B Instruct & 68.5 & 75.6 & 58.8 & 77.7 & \textbf{74.7} \\
\multirow{1}{*}{IntentRL} 
  & 4B Instruct & \textbf{82.4} & \textbf{77.5} & \textbf{74.1} & \textbf{88.7} & \textbf{74.7} \\
\bottomrule
\end{tabular}
\vspace{-5pt}
\caption{
Recall, Precision, and Full Coverage on the ambiguous and unambiguous subsets of \textbf{Ambrosia}. Best results per metric are in \textbf{bold}.
}
\label{tab:ambrosia_results}
\end{table}
\paragraph{Text-to-SQL Parsing}  Table~\ref{tab:ambrosia_results} presents our results on Ambrosia.
Again, larger models
 outperform smaller ones, while  reasoning models show no
consistent improvement over instruct models (e.g., the 4B Thinking model achieves 16.9\% recall vs.\ 22.5\% for 4B Instruct on ambiguous questions). Unlike conversational QA, models show higher
precision than recall on ambiguous questions, which means that they
often generate one answer but miss valid
interpretations. On unambiguous questions, the pattern
reverses,  indicating models struggle to calibrate when ambiguity is genuinely present.

IntentRL substantially outperforms all baselines, achieving 82.4\% recall, 77.5\% precision, and 74.1\% full coverage on ambiguous questions, while maintaining strong performance on unambiguous questions (88.7\% recall, 74.7\% precision).  High precision is particularly impressive since we do not explicitly optimise for it during training. 
There are several reasons for
this.  Firstly, the base model already shows more balanced or even
slightly higher precision than recall (e.g., 4B Instruct has~30.8\%
precision vs.\ 22.5\% recall on ambiguous questions), providing a
strong starting point. Secondly, text-to-SQL parsing allows for 
verifiable, execution-based rewards, making it easier to
learn valid interpretations compared to tasks with soft
similarity-based rewards, such as conversational QA.  
Finally, IntentRL exhibits balanced performance on ambiguous
\emph{and} unambiguous questions. It correctly identifies when
questions require multiple answers (high recall on ambiguous) and when a single answer suffices
(74.7\% precision on unambiguous). %

Appendix~\ref{appx:additional_results} provides additional results including generalization across model families.  Appendix~\ref{appx:alternative_approaches} explores alternative ambiguity-handling strategies (answer-only generation, parallel sampling, and clarification questions), finding that none outperforms our approach, though IntentRL integrates effectively with answer-only and clarification pipelines.

 \subsection{Comparison with State-of-the-art Methods}
\begin{wraptable}[8]{r}{0.29\columnwidth}
\centering
\vspace{-.4cm}
\begin{tabular}{@{}l@{\;\;\;}c@{}}
 \toprule
 \textbf{Method} & \textbf{Similarity}\\
 \midrule
 ACT & 75.1 \\
 IntentRL & 83.0 \\
 \bottomrule
\end{tabular}
\caption{Embedding-based similarity
   on \textbf{Abg-CoQA}.}
\label{tab:Abg-CoQA_act_comparison}
\end{wraptable}%
\paragraph{Conversational QA}
In Table~\ref{tab:Abg-CoQA_act_comparison}, we  compare IntentRL with ACT \citep{Chen2024LearningTC},
a state-of-the-art method that learns to disambiguate through
multi-turn clarification.
We report embedding-based semantic similarity
\citep{reimers-gurevych-2019-sentence}, macro-averaged across the test set, for direct comparison with their work (they use the term trajectory similarity).
Despite generating interpretations and corresponding answers in a single turn, IntentRL achieves a similarity of 83.0\%, significantly outperforming ACT’s 75.1\%. This  suggests that enumerating multiple  interpretations (and their answers) upfront  can be more effective than multi-turn clarification, allowing users to
identify their intended meaning immediately without additional exchanges. %

\begin{wraptable}[8]{r}{0.485\columnwidth}
 \vspace{-.4cm}
 \begin{tabular}{@{}l@{\;\;\;}c@{\;\;\;}c@{\;\;\;}c@{\;\;\;}c@{}}
  \toprule
  \small
  \textbf{Method} & \textbf{Recall} & \textbf{Prec} & \textbf{Single} & \textbf{Full} \\ 
  \midrule
  DisambigParse & 59.7 & 26.3 & 87.8 & 30.5 \\
  IntentRL & 82.4 & 77.5 & 92.4 & 74.1 \\
  \bottomrule
  \end{tabular}
  \caption{DisambigParse %
  and IntentRL results on \textbf{Ambrosia} (ambiguous questions). 
  }
  \label{tab:disambig_comparison}
\end{wraptable}
\paragraph{Text-to-SQL}  In Table~\ref{tab:disambig_comparison}, we compare IntentRL
with DisambigParse \citep{saparina-lapata-2025-disambiguate}, a 
{two-stage} method that first predicts interpretations and then generates SQL. 
In addition to our metrics, we include 
Single Coverage (percentage of questions where the model correctly
identifies \emph{at least} one valid interpretation) to match their
evaluation setup. 
We outperform DisambigParse across all metrics, achieving much higher precision (77.5\%  vs.\ 26.3\%) and  full coverage (74.1\% vs.\ 30.5\%), %
while learning from answers alone in a single stage, unlike two-stage
DisambigParse, which requires gold interpretations for training.

 \subsection{Out-of-Domain Generalization}

Following \citet{saparina-lapata-2025-disambiguate}, we evaluate on
AmbiQT \citep{bhaskar-etal-2023-benchmarking}, another text-to-SQL benchmark with ambiguous questions which serves as out-of-domain
evaluation for trainable methods. Although SFT achieves strong
in-domain performance on Ambrosia, it fails to generalise to AmbiQT, 
achieving only 9.1\% full coverage, learning dataset-specific patterns
like the structure of databases and common query templates rather than
the underlying task of identifying ambiguity. IntentRL generalises
much better, identifying all valid interpretations for nearly
half of the questions \emph{without training} on this dataset
(66.9\%~recall and 49.1\%~full coverage), 
demonstrating that IntentRL learns general reasoning skills about ambiguity that transfer across domains, databases, and queries.

\begin{table}[t]
\centering
\small
\begin{tabular}{llccc}
\toprule
\textbf{Method} & \textbf{Model} & \textbf{Recall} & \textbf{Precision} & \textbf{Full Coverage} \\
\midrule
\multirow{4}{*}{{IntentPrompt}}
  & 4B Instruct & 20.5 & 27.2 & 9.3 \\
  & 4B Thinking & 20.7 & 25.1 & 12.2 \\
  & 235B Instruct & 60.5 & 63.2 & 38.2 \\
  & 235B Thinking & 51.8 & 43.2 & 38.1 \\
\multirow{1}{*}{SFT}
  & 4B Instruct & 32.9 & 51.4 & 9.1 \\
\multirow{1}{*}{IntentRL}
  & 4B Instruct & \textbf{66.9} & 58.2 & \textbf{49.1} \\
\bottomrule
\end{tabular}
\caption{Out-of-domain evaluation on \textbf{AmbiQT} (SFT and IntentRL trained on Ambrosia).}
\label{tab:ambiqt_results}
\end{table}

 \subsection{Human Evaluation of Interpretations}\label{sec:interpretation_eval}

Automatic metrics evaluate answer quality, but do not measure whether the interpretations  genuinely  explain their answers. Abg-CoQA lacks gold interpretations, and while Ambrosia does, we did not use them during training, so our model's interpretations may differ in style and wording. We thus assess the quality of the predicted interpretations via a human study. 
We sampled 30~ambiguous examples from each dataset (80~interpretation-answer pairs in Abg-CoQA and 92 in Ambrosia),  as interpretations are most critical when multiple plausible answers exist.
For each pair, three annotators  judged whether the interpretation is compatible with  its answer (i.e.,~whether it supports or explains it). 
Annotators were given the full context, the ambiguous question, and instructed to evaluate alignment even in cases where the answer was incorrect;  an interpretable wrong answer is more useful than an unexplained one, since it allows users to quickly identify and reject it (see Appendix~\ref{app:prolific}).

\begin{table}[t]
\centering
\small
\begin{tabular}{@{}lcccc@{}}
\toprule
\textbf{Dataset} & \textbf{Alignment} & \textbf{Agreement} & $\kappa$ & \textbf{AC1} \\
\midrule
Abg-CoQA & 90.0 & 81.0 & 29.7 & 83.8 \\
Ambrosia & 91.7 & 84.0 & 27.2 & 86.5 \\
\bottomrule
\end{tabular}
\caption{
Human evaluation: alignment accuracy, inter-annotator agreement, Fleiss' $\kappa$ and Gwet's AC1.
\label{tab:interp_human_eval}}
\end{table}
Table~\ref{tab:interp_human_eval} reports average alignment accuracy across all annotations and agreement (percentage of pairs where all three annotators agree). Our model achieves 90.0\% Alignment on Abg-CoQA and 91.7\% on Ambrosia with high inter-annotator consensus (81.0--84.0\%). Fleiss' $\kappa$ is low, reflecting the well-known kappa paradox
\citep{byrt1993bias,cicchetti1990high}: over 90\% of judgments are positive, so
chance agreement  is near  87\% and deflates $\kappa$ even when observed agreement
is high. We therefore also report Gwet's AC1 \citep{gwet2008computing}, which is
robust to skewed distributions and indicates strong agreement.
These results provide a more complete picture of model performance, demonstrating that it generates meaningful interpretations that explain the provided answers. %
Aligned interpretations open up the model's decision-making, %
allowing users to select their intended meaning. %

 \subsection{Analysis of Model Output}\label{sec:analysis}
 
To understand model failures, we analyze the 30~examples from Section~\ref{sec:interpretation_eval}.
On Abg-CoQA,  we identify three types of error:  (1)~the model ignores important context from previous dialogue turns; (2)~it  sometimes produces valid interpretations but then gives factually incorrect answers, such as referencing a different person than the one implied in the interpretation; and~(3) it predicts generic interpretations like ``The question is ambiguous'', sometimes paired with the answer ``Unknown'' which   appears in some training examples as a valid gold answer, but the model has learned to produce more often than necessary.  
On Ambrosia, interpretations are sometimes vague, paraphrasing rather than disambiguating the question.
The model occasionally produces correct  interpretations but generates queries that do not follow them (e.g., adding ID fields not mentioned in the question)
 and sometimes produces 
 non-executable queries, while noting the error in SQL comments (e.g., ``this column does not exist''). 
Repeated interpretations also occur, though  these can be filtered via execution.

\begin{wrapfigure}[13]{r}{0.45\columnwidth}
\centering
\vspace{-.4cm}
    \includegraphics[width=1\linewidth]{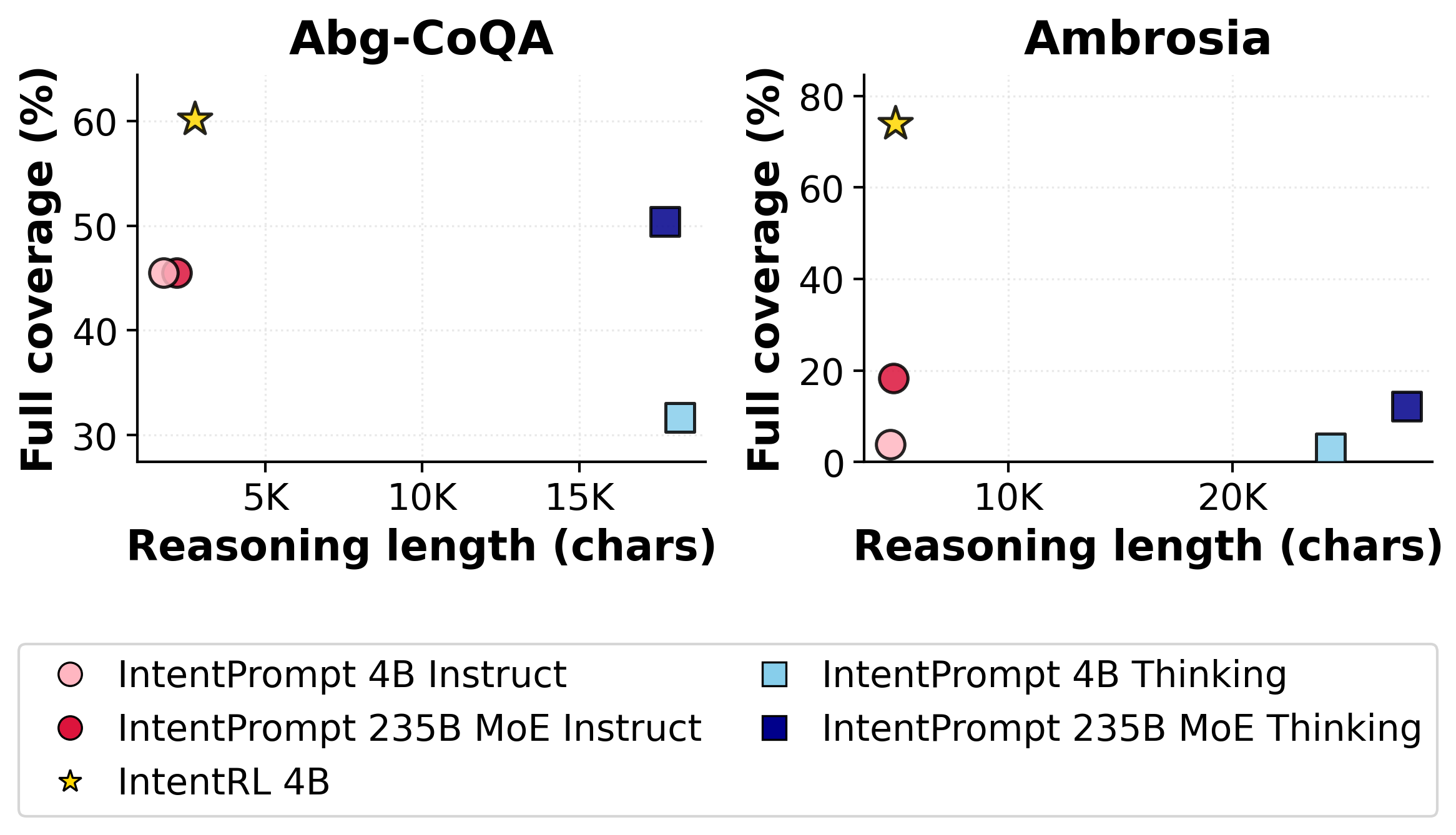}
    \caption{Reasoning length (as characters) vs.\ coverage (ambiguous subsets).}%
    \label{fig:coverage_vs_length}
\end{wrapfigure}

Next, we analyze the reasoning traces of IntentPrompt and IntentRL both quantitatively and qualitatively.
Figure~\ref{fig:coverage_vs_length} shows average reasoning length versus full coverage on ambiguous examples in Abg-CoQA and Ambrosia. 
Thinking models frequently overthink on Abg-CoQA, getting stuck on the same interpretation rather than considering alternatives, and sometimes miss interpretations entirely on Ambrosia. 
Instruct models with IntentPrompt are more concise but are outperformed by IntentRL.
Our models achieve the best balance of efficiency and coverage; they explicitly consider ambiguity and alternative interpretations before committing to an answer. 
Example traces are in Appendix~\ref{appx:model_outputs}.

\section{Conclusion}

We introduced IntentRL, a reinforcement learning approach that trains models to reason explicitly about user intent by generating structured interpretation-answer pairs for ambiguous inputs. Using a dual reward objective (recall for ambiguous inputs to maximise coverage, precision for unambiguous ones to supress spurious alternatives) our method requires only answer annotations as supervision, with no need for interpretation labels or clarification questions. Across conversational question-answering and text-to-SQL parsing, IntentRL substantially outperforms prompting and fine-tuning baselines, achieving strong gains in full coverage while maintaining competitive precision. Human evaluation confirms that predicted interpretations are highly aligned with their answers (over 90\%) despite receiving no direct supervision on this objective. Crucially, IntentRL generalises well to out-of-domain data where supervised fine-tuning fails, suggesting the model learns general reasoning about intent rather than data-specific patterns.

Beyond performance gains, our experiments help explain why the approach works. We find that ambiguity is not inherently captured in a model's output distribution: neither parallel sampling nor alternative prompting formats recover the diverse interpretations that ambiguous inputs admit, and only targeted training does. This indicates that the gains come from learning to reason about user intent rather than from the interpretation-answer format itself. Crucially, IntentRL also generalises to out-of-domain data where supervised fine-tuning collapses, suggesting that it learns transferable reasoning about intent rather than dataset-specific patterns. Where our model favours recall over precision, it does so by surfacing multiple plausible interpretations instead of silently committing to a single one. We argue this behaviour is desirable for  safety-sensitive requests.

In future work, we plan to use interpretations to detect and correct errors without regenerating the full response, and to integrate reasoning about intent into broader post-training alongside domains such as mathematics and code. Controlled decoding could further improve efficiency and accuracy, e.g., by ensuring that interpretations are mutually exclusive. Finally,  adapting to individual user preferences would make interpretations more personalized and relevant.

\section*{Acknowledgments}
We thank the anonymous reviewers and area chair for their feedback. This work was supported by the Edinburgh International Data Facility (EIDF) and the Data-Driven Innovation Programme at the University of Edinburgh. Access to EIDF was facilitated through the University of Edinburgh’s Generative AI Laboratory GAIL Fellow scheme. Lapata gratefully acknowledges the support of the UK Engineering and Physical Sciences Research Council (grant EP/W002876/1). 

\bibliography{colm2026_conference}

\begin{thebibliography}{59}
\providecommand{\natexlab}[1]{#1}
\providecommand{\url}[1]{\texttt{#1}}
\expandafter\ifx\csname urlstyle\endcsname\relax
  \providecommand{\doi}[1]{doi: #1}\else
  \providecommand{\doi}{doi: \begingroup \urlstyle{rm}\Url}\fi

\bibitem[Amplayo et~al.(2023)Amplayo, Webster, Collins, Das, and Narayan]{amplayo-etal-2023-query}
Reinald~Kim Amplayo, Kellie Webster, Michael Collins, Dipanjan Das, and Shashi Narayan.
\newblock Query refinement prompts for closed-book long-form {QA}.
\newblock In Anna Rogers, Jordan Boyd-Graber, and Naoaki Okazaki (eds.), \emph{Proceedings of the 61st Annual Meeting of the Association for Computational Linguistics (Volume 1: Long Papers)}, pp.\  7997--8012, Toronto, Canada, July 2023. Association for Computational Linguistics.
\newblock \doi{10.18653/v1/2023.acl-long.444}.
\newblock URL \url{https://aclanthology.org/2023.acl-long.444/}.

\bibitem[Berant et~al.(2025)Berant, Chen, Fisch, Aghajani, Huot, Lapata, and Eisenstein]{berant2025learning}
Jonathan Berant, Maximillian Chen, Adam Fisch, Reza Aghajani, Fantine Huot, Mirella Lapata, and Jacob Eisenstein.
\newblock Learning steerable clarification policies with collaborative self-play.
\newblock \emph{arXiv preprint arXiv:2512.04068}, 2025.

\bibitem[Bhaskar et~al.(2023)Bhaskar, Tomar, Sathe, and Sarawagi]{bhaskar-etal-2023-benchmarking}
Adithya Bhaskar, Tushar Tomar, Ashutosh Sathe, and Sunita Sarawagi.
\newblock Benchmarking and improving text-to-{SQL} generation under ambiguity.
\newblock In Houda Bouamor, Juan Pino, and Kalika Bali (eds.), \emph{Proceedings of the 2023 Conference on Empirical Methods in Natural Language Processing}, pp.\  7053--7074, Singapore, December 2023. Association for Computational Linguistics.
\newblock \doi{10.18653/v1/2023.emnlp-main.436}.
\newblock URL \url{https://aclanthology.org/2023.emnlp-main.436/}.

\bibitem[Byrt et~al.(1993)Byrt, Bishop, and Carlin]{byrt1993bias}
Ted Byrt, Janet Bishop, and John~B. Carlin.
\newblock Bias, prevalence and kappa.
\newblock \emph{Journal of Clinical Epidemiology}, 46\penalty0 (5):\penalty0 423--429, 1993.
\newblock \doi{10.1016/0895-4356(93)90018-V}.

\bibitem[Chen et~al.(2025)Chen, Sun, Arik, and Pfister]{Chen2024LearningTC}
Maximillian Chen, Ruoxi Sun, Sercan~{\"O}. Arik, and Tomas Pfister.
\newblock Learning to clarify: Multi-turn conversations with action-based contrastive self-training.
\newblock In \emph{The 13th International Conference on Learning Representations}, 2025.
\newblock URL \url{https://openreview.net/forum?id=SIE6VFps9x}.

\bibitem[Choi et~al.(2025)Choi, Bansal, and Stengel-Eskin]{choi2025language}
Jio Choi, Mohit Bansal, and Elias Stengel-Eskin.
\newblock Language models identify ambiguities and exploit loopholes.
\newblock \emph{arXiv preprint arXiv:2508.19546}, 2025.

\bibitem[Cicchetti \& Feinstein(1990)Cicchetti and Feinstein]{cicchetti1990high}
Domenic~V. Cicchetti and Alvan~R. Feinstein.
\newblock High agreement but low kappa: {II}. resolving the paradoxes.
\newblock \emph{Journal of Clinical Epidemiology}, 43\penalty0 (6):\penalty0 551--558, 1990.
\newblock \doi{10.1016/0895-4356(90)90159-M}.

\bibitem[Cobbe et~al.(2021)Cobbe, Kosaraju, Bavarian, Chen, Jun, Kaiser, Plappert, Tworek, Hilton, Nakano, Hesse, and Schulman]{cobbe2021gsm8k}
Karl Cobbe, Vineet Kosaraju, Mohammad Bavarian, Mark Chen, Heewoo Jun, Lukasz Kaiser, Matthias Plappert, Jerry Tworek, Jacob Hilton, Reiichiro Nakano, Christopher Hesse, and John Schulman.
\newblock Training verifiers to solve math word problems.
\newblock \emph{arXiv preprint arXiv:2110.14168}, 2021.

\bibitem[Dao(2024)]{dao2023flashattention2}
Tri Dao.
\newblock Flash{A}ttention-2: Faster attention with better parallelism and work partitioning.
\newblock In \emph{International Conference on Learning Representations (ICLR)}, 2024.

\bibitem[Deng et~al.(2022)Deng, Lei, Zhang, Lam, and Chua]{emnlp22-pacific}
Yang Deng, Wenqiang Lei, Wenxuan Zhang, Wai Lam, and Tat{-}Seng Chua.
\newblock {PACIFIC:} towards proactive conversational question answering over tabular and textual data in finance.
\newblock In \emph{Proceedings of the 2022 Conference on Empirical Methods in Natural Language Processing, {EMNLP} 2022}, 2022.

\bibitem[Ding et~al.(2025)Ding, Lin, and Zeng]{ding2025ambisql}
Zhongjun Ding, Yin Lin, and Tianjing Zeng.
\newblock Ambisql: Interactive ambiguity detection and resolution for text-to-sql.
\newblock \emph{arXiv preprint arXiv:2508.15276}, 2025.

\bibitem[Dong et~al.(2025)Dong, Ashok~Kumar, Hu, Chauhan, Hang, Chang, Pan, Lan, Zhu, Jiang, Ng, and Wang]{dong-etal-2025-practiq}
Mingwen Dong, Nischal Ashok~Kumar, Yiqun Hu, Anuj Chauhan, Chung-Wei Hang, Shuaichen Chang, Lin Pan, Wuwei Lan, Henghui Zhu, Jiarong Jiang, Patrick Ng, and Zhiguo Wang.
\newblock {PRACTIQ}: A practical conversational text-to-{SQL} dataset with ambiguous and unanswerable queries.
\newblock In Luis Chiruzzo, Alan Ritter, and Lu~Wang (eds.), \emph{Proceedings of the 2025 Conference of the Nations of the Americas Chapter of the Association for Computational Linguistics: Human Language Technologies (Volume 1: Long Papers)}, pp.\  255--273, Albuquerque, New Mexico, April 2025. Association for Computational Linguistics.
\newblock ISBN 979-8-89176-189-6.
\newblock \doi{10.18653/v1/2025.naacl-long.13}.
\newblock URL \url{https://aclanthology.org/2025.naacl-long.13/}.

\bibitem[Floratou et~al.(2024)Floratou, Psallidas, Zhao, Deep, Hagleither, Tan, Cahoon, Alotaibi, Henkel, Singla, Grootel, Chow, Deng, Lin, Campos, Emani, Pandit, Shnayder, Wang, and Curino]{Floratou2024}
Avrilia Floratou, Fotis Psallidas, Fuheng Zhao, Shaleen Deep, Gunther Hagleither, Wangda Tan, Joyce Cahoon, Rana Alotaibi, Jordan Henkel, Abhik Singla, Alex~Van Grootel, Brandon Chow, Kai Deng, Katherine Lin, Marcos Campos, K.~Venkatesh Emani, Vivek Pandit, Victor Shnayder, Wenjing Wang, and Carlo Curino.
\newblock {NL2SQL} is a solved problem... not!
\newblock In \emph{14th Conference on Innovative Data Systems Research, {CIDR}}, 2024.
\newblock URL \url{https://www.cidrdb.org/cidr2024/papers/p74-floratou.pdf}.

\bibitem[Gema et~al.(2024)Gema, Leang, Hong, Devoto, Mancino, Saxena, He, Zhao, Du, Madani, Barale, McHardy, Harris, Kaddour, van Krieken, and Minervini]{gema2024mmlu}
Aryo~Pradipta Gema, Joshua Ong~Jun Leang, Giwon Hong, Alessio Devoto, Alberto Carlo~Maria Mancino, Rohit Saxena, Xuanli He, Yu~Zhao, Xiaotang Du, Mohammad Reza~Ghasemi Madani, Claire Barale, Robert McHardy, Joshua Harris, Jean Kaddour, Emile van Krieken, and Pasquale Minervini.
\newblock Are we done with mmlu?, 2024.

\bibitem[Guo et~al.(2021)Guo, Zhang, Reddy, and Alikhani]{guo2021AbgCoQA}
Meiqi Guo, Mingda Zhang, Siva Reddy, and Malihe Alikhani.
\newblock Abg-co{QA}: Clarifying ambiguity in conversational question answering.
\newblock In \emph{3rd Conference on Automated Knowledge Base Construction}, 2021.
\newblock URL \url{https://openreview.net/forum?id=SlDZ1o8FsJU}.

\bibitem[Gwet(2008)]{gwet2008computing}
Kilem~L. Gwet.
\newblock Computing inter-rater reliability and its variance in the presence of high agreement.
\newblock \emph{British Journal of Mathematical and Statistical Psychology}, 61\penalty0 (1):\penalty0 29--48, 2008.
\newblock \doi{10.1348/000711006X126600}.

\bibitem[Hendrycks et~al.(2021)Hendrycks, Burns, Basart, Zou, Mazeika, Song, and Steinhardt]{hendryckstest2021}
Dan Hendrycks, Collin Burns, Steven Basart, Andy Zou, Mantas Mazeika, Dawn Song, and Jacob Steinhardt.
\newblock Measuring massive multitask language understanding.
\newblock \emph{Proceedings of the International Conference on Learning Representations (ICLR)}, 2021.

\bibitem[Hui et~al.(2024)Hui, Yang, Cui, Yang, Liu, Zhang, Liu, Zhang, Yu, Lu, Dang, Fan, Zhang, Yang, Men, Huang, Zheng, Miao, Quan, Feng, Ren, Ren, Zhou, and Lin]{hui2024qwen25codertechnicalreport}
Binyuan Hui, Jian Yang, Zeyu Cui, Jiaxi Yang, Dayiheng Liu, Lei Zhang, Tianyu Liu, Jiajun Zhang, Bowen Yu, Keming Lu, Kai Dang, Yang Fan, Yichang Zhang, An~Yang, Rui Men, Fei Huang, Bo~Zheng, Yibo Miao, Shanghaoran Quan, Yunlong Feng, Xingzhang Ren, Xuancheng Ren, Jingren Zhou, and Junyang Lin.
\newblock Qwen2.5-coder technical report, 2024.
\newblock URL \url{https://arxiv.org/abs/2409.12186}.

\bibitem[Huo et~al.(2025)Huo, Xu, Li, Jacobsson, Lin, Qin, Hui, Li, Qu, Si, et~al.]{huo2025bird}
Nan Huo, Xiaohan Xu, Jinyang Li, Per Jacobsson, Shipei Lin, Bowen Qin, Binyuan Hui, Xiaolong Li, Ge~Qu, Shuzheng Si, et~al.
\newblock Bird-interact: Re-imagining text-to-sql evaluation for large language models via lens of dynamic interactions.
\newblock \emph{arXiv preprint arXiv:2510.05318}, 2025.

\bibitem[In et~al.(2025)In, Kim, Rossi, Tanjim, Yu, Sinha, and Park]{in-etal-2025-diversify}
Yeonjun In, Sungchul Kim, Ryan~A. Rossi, Mehrab Tanjim, Tong Yu, Ritwik Sinha, and Chanyoung Park.
\newblock Diversify-verify-adapt: Efficient and robust retrieval-augmented ambiguous question answering.
\newblock In Luis Chiruzzo, Alan Ritter, and Lu~Wang (eds.), \emph{Proceedings of the 2025 Conference of the Nations of the Americas Chapter of the Association for Computational Linguistics: Human Language Technologies (Volume 1: Long Papers)}, pp.\  1212--1233, Albuquerque, New Mexico, April 2025. Association for Computational Linguistics.
\newblock ISBN 979-8-89176-189-6.
\newblock \doi{10.18653/v1/2025.naacl-long.56}.
\newblock URL \url{https://aclanthology.org/2025.naacl-long.56/}.

\bibitem[Kobalczyk et~al.(2025)Kobalczyk, Astorga, Liu, and van~der Schaar]{kobalczyk2025active}
Kasia Kobalczyk, Nicol{\'a}s Astorga, Tennison Liu, and Mihaela van~der Schaar.
\newblock Active task disambiguation with {LLM}s.
\newblock In \emph{The Thirteenth International Conference on Learning Representations}, 2025.
\newblock URL \url{https://openreview.net/forum?id=JAMxRSXLFz}.

\bibitem[Kwiatkowski et~al.(2019)Kwiatkowski, Palomaki, Redfield, Collins, Parikh, Alberti, Epstein, Polosukhin, Devlin, Lee, Toutanova, Jones, Kelcey, Chang, Dai, Uszkoreit, Le, and Petrov]{kwiatkowski-etal-2019-natural}
Tom Kwiatkowski, Jennimaria Palomaki, Olivia Redfield, Michael Collins, Ankur Parikh, Chris Alberti, Danielle Epstein, Illia Polosukhin, Jacob Devlin, Kenton Lee, Kristina Toutanova, Llion Jones, Matthew Kelcey, Ming-Wei Chang, Andrew~M. Dai, Jakob Uszkoreit, Quoc Le, and Slav Petrov.
\newblock Natural questions: A benchmark for question answering research.
\newblock \emph{Transactions of the Association for Computational Linguistics}, 7:\penalty0 452--466, 2019.
\newblock \doi{10.1162/tacl_a_00276}.
\newblock URL \url{https://aclanthology.org/Q19-1026/}.

\bibitem[Kwon et~al.(2023)Kwon, Li, Zhuang, Sheng, Zheng, Yu, Gonzalez, Zhang, and Stoica]{kwon2023efficient}
Woosuk Kwon, Zhuohan Li, Siyuan Zhuang, Ying Sheng, Lianmin Zheng, Cody~Hao Yu, Joseph~E. Gonzalez, Hao Zhang, and Ion Stoica.
\newblock Efficient memory management for large language model serving with pagedattention.
\newblock In \emph{Proceedings of the ACM SIGOPS 29th Symposium on Operating Systems Principles}, 2023.

\bibitem[Li et~al.(2023)Li, Mesgar, Martins, and Gurevych]{li-etal-2023-python}
Haau-Sing~(Xiaocheng) Li, Mohsen Mesgar, Andr{\'e} Martins, and Iryna Gurevych.
\newblock Python code generation by asking clarification questions.
\newblock In Anna Rogers, Jordan Boyd-Graber, and Naoaki Okazaki (eds.), \emph{Proceedings of the 61st Annual Meeting of the Association for Computational Linguistics (Volume 1: Long Papers)}, pp.\  14287--14306, Toronto, Canada, July 2023. Association for Computational Linguistics.
\newblock \doi{10.18653/v1/2023.acl-long.799}.
\newblock URL \url{https://aclanthology.org/2023.acl-long.799/}.

\bibitem[Liu et~al.(2025)Liu, Chen, Li, Qi, Pang, Du, Lee, and Lin]{liu2025understanding}
Zichen Liu, Changyu Chen, Wenjun Li, Penghui Qi, Tianyu Pang, Chao Du, Wee~Sun Lee, and Min Lin.
\newblock Understanding r1-zero-like training: A critical perspective.
\newblock In \emph{Second Conference on Language Modeling}, 2025.
\newblock URL \url{https://openreview.net/forum?id=5PAF7PAY2Y}.

\bibitem[Min et~al.(2020)Min, Michael, Hajishirzi, and Zettlemoyer]{min-etal-2020-ambigqa}
Sewon Min, Julian Michael, Hannaneh Hajishirzi, and Luke Zettlemoyer.
\newblock {A}mbig{QA}: Answering ambiguous open-domain questions.
\newblock In Bonnie Webber, Trevor Cohn, Yulan He, and Yang Liu (eds.), \emph{Proceedings of the 2020 Conference on Empirical Methods in Natural Language Processing (EMNLP)}, pp.\  5783--5797, Online, November 2020. Association for Computational Linguistics.
\newblock \doi{10.18653/v1/2020.emnlp-main.466}.
\newblock URL \url{https://aclanthology.org/2020.emnlp-main.466/}.

\bibitem[Min et~al.(2021)Min, Lee, Chang, Toutanova, and Hajishirzi]{min-etal-2021-joint}
Sewon Min, Kenton Lee, Ming-Wei Chang, Kristina Toutanova, and Hannaneh Hajishirzi.
\newblock Joint passage ranking for diverse multi-answer retrieval.
\newblock In Marie-Francine Moens, Xuanjing Huang, Lucia Specia, and Scott Wen-tau Yih (eds.), \emph{Proceedings of the 2021 Conference on Empirical Methods in Natural Language Processing}, pp.\  6997--7008, Online and Punta Cana, Dominican Republic, November 2021. Association for Computational Linguistics.
\newblock \doi{10.18653/v1/2021.emnlp-main.560}.
\newblock URL \url{https://aclanthology.org/2021.emnlp-main.560/}.

\bibitem[Mu et~al.(2024)Mu, Shi, Wang, Yu, Zhang, Wang, Liu, and Wang]{mu2023clarifygpt}
Fangwen Mu, Lin Shi, Song Wang, Zhuohao Yu, Binquan Zhang, Chenxue Wang, Shichao Liu, and Qing Wang.
\newblock Clarifygpt: Empowering llm-based code generation with intention clarification, 2024.
\newblock URL \url{https://arxiv.org/pdf/2310.10996}.

\bibitem[Pei et~al.(2022)Pei, Ananthasubramaniam, Wang, Zhou, Dedeloudis, Sargent, and Jurgens]{pei-etal-2022-potato}
Jiaxin Pei, Aparna Ananthasubramaniam, Xingyao Wang, Naitian Zhou, Apostolos Dedeloudis, Jackson Sargent, and David Jurgens.
\newblock {POTATO}: The portable text annotation tool.
\newblock In Wanxiang Che and Ekaterina Shutova (eds.), \emph{Proceedings of the 2022 Conference on Empirical Methods in Natural Language Processing: System Demonstrations}, pp.\  327--337, Abu Dhabi, UAE, December 2022. Association for Computational Linguistics.
\newblock \doi{10.18653/v1/2022.emnlp-demos.33}.
\newblock URL \url{https://aclanthology.org/2022.emnlp-demos.33/}.

\bibitem[Piantadosi et~al.(2012)Piantadosi, Tily, and Gibson]{piantadosi2012communicative}
Steven~T Piantadosi, Harry Tily, and Edward Gibson.
\newblock The communicative function of ambiguity in language.
\newblock \emph{Cognition}, 122\penalty0 (3):\penalty0 280--291, 2012.

\bibitem[Pourreza \& Rafiei(2023)Pourreza and Rafiei]{pourreza-rafiei-2023-evaluating}
Mohammadreza Pourreza and Davood Rafiei.
\newblock Evaluating cross-domain text-to-{SQL} models and benchmarks.
\newblock In Houda Bouamor, Juan Pino, and Kalika Bali (eds.), \emph{Proceedings of the 2023 Conference on Empirical Methods in Natural Language Processing}, pp.\  1601--1611, Singapore, December 2023. Association for Computational Linguistics.
\newblock \doi{10.18653/v1/2023.emnlp-main.99}.
\newblock URL \url{https://aclanthology.org/2023.emnlp-main.99/}.

\bibitem[Puri et~al.(2026)Puri, Damani, Shenfeld, Ghassemi, Andreas, and Kim]{puri2026reaching}
Isha Puri, Mehul Damani, Idan Shenfeld, Marzyeh Ghassemi, Jacob Andreas, and Yoon Kim.
\newblock Reaching beyond the mode: Rl for distributional reasoning in language models.
\newblock \emph{arXiv preprint arXiv:2603.24844}, 2026.

\bibitem[Rasley et~al.(2020)Rasley, Rajbhandari, Ruwase, and He]{deepspeed}
Jeff Rasley, Samyam Rajbhandari, Olatunji Ruwase, and Yuxiong He.
\newblock Deepspeed: System optimizations enable training deep learning models with over 100 billion parameters.
\newblock In \emph{Proceedings of the 26th ACM SIGKDD International Conference on Knowledge Discovery \& Data Mining}, KDD '20, pp.\  3505–3506, New York, NY, USA, 2020. Association for Computing Machinery.
\newblock ISBN 9781450379984.
\newblock \doi{10.1145/3394486.3406703}.
\newblock URL \url{https://doi.org/10.1145/3394486.3406703}.

\bibitem[Rasmussen \& Schuler(2020)Rasmussen and Schuler]{rasmussen-schuler-2020-corpus}
Nathan Rasmussen and William Schuler.
\newblock A corpus of encyclopedia articles with logical forms.
\newblock In Nicoletta Calzolari, Fr{\'e}d{\'e}ric B{\'e}chet, Philippe Blache, Khalid Choukri, Christopher Cieri, Thierry Declerck, Sara Goggi, Hitoshi Isahara, Bente Maegaard, Joseph Mariani, H{\'e}l{\`e}ne Mazo, Asuncion Moreno, Jan Odijk, and Stelios Piperidis (eds.), \emph{Proceedings of the Twelfth Language Resources and Evaluation Conference}, pp.\  1051--1060, Marseille, France, May 2020. European Language Resources Association.
\newblock ISBN 979-10-95546-34-4.
\newblock URL \url{https://aclanthology.org/2020.lrec-1.132/}.

\bibitem[Reimers \& Gurevych(2019)Reimers and Gurevych]{reimers-gurevych-2019-sentence}
Nils Reimers and Iryna Gurevych.
\newblock Sentence-{BERT}: Sentence embeddings using {S}iamese {BERT}-networks.
\newblock In Kentaro Inui, Jing Jiang, Vincent Ng, and Xiaojun Wan (eds.), \emph{Proceedings of the 2019 Conference on Empirical Methods in Natural Language Processing and the 9th International Joint Conference on Natural Language Processing (EMNLP-IJCNLP)}, pp.\  3982--3992, Hong Kong, China, November 2019. Association for Computational Linguistics.
\newblock \doi{10.18653/v1/D19-1410}.
\newblock URL \url{https://aclanthology.org/D19-1410/}.

\bibitem[Saparina \& Lapata(2024)Saparina and Lapata]{ambrosia}
Irina Saparina and Mirella Lapata.
\newblock Ambrosia: A benchmark for parsing ambiguous questions into database queries.
\newblock \emph{Proceedings of the 38th Annual Conference on Neural Information Processing Systems}, 2024.
\newblock URL \url{https://proceedings.neurips.cc/paper_files/paper/2024/hash/a4c942a8405cc910f0a833d28d2573cc-Abstract-Datasets_and_Benchmarks_Track.html}.

\bibitem[Saparina \& Lapata(2025)Saparina and Lapata]{saparina-lapata-2025-disambiguate}
Irina Saparina and Mirella Lapata.
\newblock Disambiguate first, parse later: Generating interpretations for ambiguity resolution in semantic parsing.
\newblock In Wanxiang Che, Joyce Nabende, Ekaterina Shutova, and Mohammad~Taher Pilehvar (eds.), \emph{Findings of the Association for Computational Linguistics: ACL 2025}, pp.\  16825--16839, Vienna, Austria, July 2025. Association for Computational Linguistics.
\newblock ISBN 979-8-89176-256-5.
\newblock \doi{10.18653/v1/2025.findings-acl.863}.
\newblock URL \url{https://aclanthology.org/2025.findings-acl.863/}.

\bibitem[Shao et~al.(2024)Shao, Wang, Zhu, Xu, Song, Bi, Zhang, Zhang, Li, Wu, and Guo]{deepseek-math}
Zhihong Shao, Peiyi Wang, Qihao Zhu, Runxin Xu, Junxiao Song, Xiao Bi, Haowei Zhang, Mingchuan Zhang, Y.~K. Li, Y.~Wu, and Daya Guo.
\newblock Deepseekmath: Pushing the limits of mathematical reasoning in open language models, 2024.
\newblock URL \url{https://arxiv.org/abs/2402.03300}.

\bibitem[Sorensen et~al.(2024)Sorensen, Moore, Fisher, Gordon, Mireshghallah, Rytting, Ye, Jiang, Lu, Dziri, Althoff, and Choi]{sorensen2024}
Taylor Sorensen, Jared Moore, Jillian Fisher, Mitchell Gordon, Niloofar Mireshghallah, Christopher~Michael Rytting, Andre Ye, Liwei Jiang, Ximing Lu, Nouha Dziri, Tim Althoff, and Yejin Choi.
\newblock Position: a roadmap to pluralistic alignment.
\newblock In \emph{Proceedings of the 41st International Conference on Machine Learning}, ICML'24. JMLR.org, 2024.

\bibitem[Sorensen et~al.(2025)Sorensen, Newman, Moore, Park, Fisher, Mireshghallah, Jiang, and Choi]{sorensen2025spectrumtuningposttrainingdistributional}
Taylor Sorensen, Benjamin Newman, Jared Moore, Chan Park, Jillian Fisher, Niloofar Mireshghallah, Liwei Jiang, and Yejin Choi.
\newblock Spectrum tuning: Post-training for distributional coverage and in-context steerability, 2025.
\newblock URL \url{https://arxiv.org/abs/2510.06084}.

\bibitem[Stelmakh et~al.(2022)Stelmakh, Luan, Dhingra, and Chang]{stelmakh-etal-2022-asqa}
Ivan Stelmakh, Yi~Luan, Bhuwan Dhingra, and Ming-Wei Chang.
\newblock {ASQA}: Factoid questions meet long-form answers.
\newblock In Yoav Goldberg, Zornitsa Kozareva, and Yue Zhang (eds.), \emph{Proceedings of the 2022 Conference on Empirical Methods in Natural Language Processing}, pp.\  8273--8288, Abu Dhabi, United Arab Emirates, December 2022. Association for Computational Linguistics.
\newblock \doi{10.18653/v1/2022.emnlp-main.566}.
\newblock URL \url{https://aclanthology.org/2022.emnlp-main.566/}.

\bibitem[Stengel-Eskin et~al.(2024)Stengel-Eskin, Rawlins, and Durme]{eskin-2024-zero}
Elias Stengel-Eskin, Kyle Rawlins, and Benjamin~Van Durme.
\newblock Zero and few-shot semantic parsing with ambiguous inputs.
\newblock In \emph{The 12th International Conference on Learning Representations}, 2024.
\newblock URL \url{https://openreview.net/forum?id=qL9gogRepu}.

\bibitem[Sun et~al.(2024)Sun, Xu, Zha, Liu, and Dong]{sun-etal-2024-head}
Kai Sun, Yifan Xu, Hanwen Zha, Yue Liu, and Xin~Luna Dong.
\newblock Head-to-tail: How knowledgeable are large language models ({LLM}s)? {A}.{K}.{A}. will {LLM}s replace knowledge graphs?
\newblock In Kevin Duh, Helena Gomez, and Steven Bethard (eds.), \emph{Proceedings of the 2024 Conference of the North American Chapter of the Association for Computational Linguistics: Human Language Technologies (Volume 1: Long Papers)}, pp.\  311--325, Mexico City, Mexico, June 2024. Association for Computational Linguistics.
\newblock \doi{10.18653/v1/2024.naacl-long.18}.
\newblock URL \url{https://aclanthology.org/2024.naacl-long.18/}.

\bibitem[Tanjim et~al.(2025)Tanjim, In, Chen, Bursztyn, Rossi, Kim, Ren, Muppala, Jiang, Kim, and Park]{tanjim2025disambiguation}
Mehrab Tanjim, Yeonjun In, Xiang Chen, Victor Bursztyn, Ryan~A. Rossi, Sungchul Kim, Guang-Jie Ren, Vaishnavi Muppala, Shun Jiang, Yongsung Kim, and Chanyoung Park.
\newblock Disambiguation in conversational question answering in the era of {LLM}s and agents: A survey.
\newblock In Christos Christodoulopoulos, Tanmoy Chakraborty, Carolyn Rose, and Violet Peng (eds.), \emph{Proceedings of the 2025 Conference on Empirical Methods in Natural Language Processing}, pp.\  9537--9550, Suzhou, China, November 2025. Association for Computational Linguistics.
\newblock ISBN 979-8-89176-332-6.
\newblock \doi{10.18653/v1/2025.emnlp-main.482}.
\newblock URL \url{https://aclanthology.org/2025.emnlp-main.482/}.

\bibitem[Team et~al.(2025)Team, Anil, Borgeaud, Alayrac, Yu, Soricut, Schalkwyk, Dai, Hauth, et~al.]{geminiteam2025geminifamilyhighlycapable}
Gemini Team, Rohan Anil, Sebastian Borgeaud, Jean-Baptiste Alayrac, Jiahui Yu, Radu Soricut, Johan Schalkwyk, Andrew~M. Dai, Anja Hauth, et~al.
\newblock Gemini: A family of highly capable multimodal models, 2025.
\newblock URL \url{https://arxiv.org/abs/2312.11805}.

\bibitem[Team et~al.(2024)Team, Dubey, Jauhri, Pandey, Kadian, Al-Dahle, Letman, Mathur, Schelten, Yang, Fan, Goyal, Hartshorn, Yang, Mitra, Sravankumar, Korenev, Hinsvark, Rao, Zhang, Rodriguez, Gregerson, Spataru, tiste Roziere, Biron, Tang, et~al.]{Dubey2024TheL3}
Llama~3 Team, Abhimanyu Dubey, Abhinav Jauhri, Abhinav Pandey, Abhishek Kadian, Ahmad Al-Dahle, Aiesha Letman, Akhil Mathur, Alan Schelten, Amy Yang, Angela Fan, Anirudh Goyal, Anthony~S. Hartshorn, Aobo Yang, Archi Mitra, Archie Sravankumar, Artem Korenev, Arthur Hinsvark, Arun Rao, Aston Zhang, Aur{\'e}lien Rodriguez, Austen Gregerson, Ava Spataru, Bap tiste Roziere, Bethany Biron, Binh Tang, et~al.
\newblock The llama 3 herd of models.
\newblock \emph{ArXiv}, abs/2407.21783, 2024.
\newblock URL \url{https://api.semanticscholar.org/CorpusID:271571434}.

\bibitem[Troshin et~al.(2025)Troshin, Saparina, Fokkens, and Niculae]{troshin2025asking}
Sergey Troshin, Irina Saparina, Antske Fokkens, and Vlad Niculae.
\newblock Asking a language model for diverse responses.
\newblock \emph{arXiv preprint arXiv:2509.17570}, 2025.

\bibitem[Tunstall et~al.(2023)Tunstall, Beeching, Lambert, Rajani, Rasul, Belkada, Huang, von Werra, Fourrier, Habib, Sarrazin, Sanseviero, Rush, and Wolf]{tunstall2023zephyrdirectdistillationlm}
Lewis Tunstall, Edward Beeching, Nathan Lambert, Nazneen Rajani, Kashif Rasul, Younes Belkada, Shengyi Huang, Leandro von Werra, Clémentine Fourrier, Nathan Habib, Nathan Sarrazin, Omar Sanseviero, Alexander~M. Rush, and Thomas Wolf.
\newblock Zephyr: Direct distillation of lm alignment, 2023.
\newblock URL \url{https://arxiv.org/abs/2310.16944}.

\bibitem[von Werra et~al.(2020)von Werra, Belkada, Tunstall, Beeching, Thrush, Lambert, Huang, Rasul, and Gallouédec]{vonwerra2022trl}
Leandro von Werra, Younes Belkada, Lewis Tunstall, Edward Beeching, Tristan Thrush, Nathan Lambert, Shengyi Huang, Kashif Rasul, and Quentin Gallouédec.
\newblock Trl: Transformer reinforcement learning.
\newblock \url{https://github.com/huggingface/trl}, 2020.

\bibitem[Wang et~al.(2023)Wang, Gao, Li, and Lou]{wang-etal-2023-know}
Bing Wang, Yan Gao, Zhoujun Li, and Jian-Guang Lou.
\newblock Know what {I} don{'}t know: Handling ambiguous and unknown questions for text-to-{SQL}.
\newblock In Anna Rogers, Jordan Boyd-Graber, and Naoaki Okazaki (eds.), \emph{Findings of the Association for Computational Linguistics: ACL 2023}, pp.\  5701--5714, Toronto, Canada, July 2023. Association for Computational Linguistics.
\newblock \doi{10.18653/v1/2023.findings-acl.352}.
\newblock URL \url{https://aclanthology.org/2023.findings-acl.352/}.

\bibitem[Wang et~al.(2024)Wang, Ma, Zhang, Ni, Chandra, Guo, Ren, Arulraj, He, Jiang, et~al.]{wang2024mmlu}
Yubo Wang, Xueguang Ma, Ge~Zhang, Yuansheng Ni, Abhranil Chandra, Shiguang Guo, Weiming Ren, Aaran Arulraj, Xuan He, Ziyan Jiang, et~al.
\newblock Mmlu-pro: A more robust and challenging multi-task language understanding benchmark.
\newblock \emph{arXiv preprint arXiv:2406.01574}, 2024.

\bibitem[West \& Potts(2025)West and Potts]{west2025base}
Peter West and Christopher Potts.
\newblock Base models beat aligned models at randomness and creativity.
\newblock In \emph{Second Conference on Language Modeling}, 2025.
\newblock URL \url{https://openreview.net/forum?id=vqN8uom4A1}.

\bibitem[Yang et~al.(2025)Yang, Li, Yang, Zhang, Hui, Zheng, Yu, Gao, Huang, Lv, Zheng, Liu, Zhou, Huang, Hu, Ge, Wei, Lin, Tang, Yang, Tu, Zhang, Yang, Yang, Zhou, Zhou, Lin, Dang, Bao, Yang, Yu, Deng, Li, Xue, Li, Zhang, Wang, Zhu, Men, Gao, Liu, Luo, Li, Tang, Yin, Ren, Wang, Zhang, Ren, Fan, Su, Zhang, Zhang, Wan, Liu, Wang, Cui, Zhang, Zhou, and Qiu]{yang2025qwen3technicalreport}
An~Yang, Anfeng Li, Baosong Yang, Beichen Zhang, Binyuan Hui, Bo~Zheng, Bowen Yu, Chang Gao, Chengen Huang, Chenxu Lv, Chujie Zheng, Dayiheng Liu, Fan Zhou, Fei Huang, Feng Hu, Hao Ge, Haoran Wei, Huan Lin, Jialong Tang, Jian Yang, Jianhong Tu, Jianwei Zhang, Jianxin Yang, Jiaxi Yang, Jing Zhou, Jingren Zhou, Junyang Lin, Kai Dang, Keqin Bao, Kexin Yang, Le~Yu, Lianghao Deng, Mei Li, Mingfeng Xue, Mingze Li, Pei Zhang, Peng Wang, Qin Zhu, Rui Men, Ruize Gao, Shixuan Liu, Shuang Luo, Tianhao Li, Tianyi Tang, Wenbiao Yin, Xingzhang Ren, Xinyu Wang, Xinyu Zhang, Xuancheng Ren, Yang Fan, Yang Su, Yichang Zhang, Yinger Zhang, Yu~Wan, Yuqiong Liu, Zekun Wang, Zeyu Cui, Zhenru Zhang, Zhipeng Zhou, and Zihan Qiu.
\newblock Qwen3 technical report, 2025.
\newblock URL \url{https://arxiv.org/abs/2505.09388}.

\bibitem[Yu et~al.(2025)Yu, Zhang, Zhu, Yuan, Zuo, Yue, Dai, Fan, Liu, Liu, et~al.]{yu2025dapo}
Qiying Yu, Zheng Zhang, Ruofei Zhu, Yufeng Yuan, Xiaochen Zuo, Yu~Yue, Weinan Dai, Tiantian Fan, Gaohong Liu, Lingjun Liu, et~al.
\newblock Dapo: An open-source llm reinforcement learning system at scale.
\newblock \emph{arXiv preprint arXiv:2503.14476}, 2025.

\bibitem[Zhang et~al.(2025{\natexlab{a}})Zhang, Yu, Chong, Sicilia, Tomz, Manning, and Shi]{zhang2025verbalized}
Jiayi Zhang, Simon Yu, Derek Chong, Anthony Sicilia, Michael~R Tomz, Christopher~D Manning, and Weiyan Shi.
\newblock Verbalized sampling: How to mitigate mode collapse and unlock llm diversity.
\newblock \emph{arXiv preprint arXiv:2510.01171}, 2025{\natexlab{a}}.

\bibitem[Zhang et~al.(2025{\natexlab{b}})Zhang, Knox, and Choi]{zhang2025modeling}
Michael~JQ Zhang, W.~Bradley Knox, and Eunsol Choi.
\newblock Modeling future conversation turns to teach {LLM}s to ask clarifying questions.
\newblock In \emph{The Thirteenth International Conference on Learning Representations}, 2025{\natexlab{b}}.
\newblock URL \url{https://openreview.net/forum?id=cwuSAR7EKd}.

\bibitem[Zhang et~al.(2024)Zhang, Schwarzschild, Carlini, Kolter, and Ippolito]{zhang2024forcing}
Yiming Zhang, Avi Schwarzschild, Nicholas Carlini, J~Zico Kolter, and Daphne Ippolito.
\newblock Forcing diffuse distributions out of language models.
\newblock In \emph{First Conference on Language Modeling}, 2024.
\newblock URL \url{https://openreview.net/forum?id=9JY1QLVFPZ}.

\bibitem[Zhang et~al.(2025{\natexlab{c}})Zhang, Diddee, Holm, Liu, Liu, Samuel, Wang, and Ippolito]{zhang2025noveltybench}
Yiming Zhang, Harshita Diddee, Susan Holm, Hanchen Liu, Xinyue Liu, Vinay Samuel, Barry Wang, and Daphne Ippolito.
\newblock Noveltybench: Evaluating creativity and diversity in language models.
\newblock In \emph{Second Conference on Language Modeling}, 2025{\natexlab{c}}.
\newblock URL \url{https://openreview.net/forum?id=XZm1ekzERf}.

\bibitem[Zipf(1949)]{zipf.g.1949}
George~Kingsley Zipf.
\newblock \emph{Human Behaviour and the Principle of Least Effort}.
\newblock Addison-Wesley Press, 1949.

\end{thebibliography}
\bibliographystyle{colm2026_conference}

\newpage
\appendix
\section{IntentPrompt}\label{appx:prompts}
Below we provide the prompt template for reasoning about user intent with explicit interpretations, used in IntentPrompt baseline and IntentRL training:

\begin{tcolorbox}[
  colback=white,
  colframe=prompt_colour,
  colbacktitle=prompt_colour,
  coltitle=white,
  title=\textbf{IntentPrompt Template}, breakable, halign=flush left
]
    \fontsize{8}{9.5}\selectfont
      You receive a context and a question. First analyse the question to identify different possible interpretations, then provide the corresponding answer for each interpretation.\\\vspace{2.75mm}
  If the question is unambiguous, provide a single clear interpretation and answer. If the question contains ambiguities, underspecifications, or could be understood in multiple ways, provide multiple plausible interpretations (maximum of 5) and corresponding answers.\\\vspace{2.75mm}
  An interpretation is a self-contained, precise reformulation of the question that clarifies scope, entities, and intent so that the question could be answered unambiguously. Different interpretations should lead to meaningfully different answers.\\\vspace{2.75mm}
  Context:\\
  \texttt{\textcolor{prompt_colour}{\{context\}}}\\\vspace{2.75mm}
  Question to interpret and answer:\\
  \texttt{\textcolor{prompt_colour}{\{question\}}}\\\vspace{2.75mm}
  General Instructions:\\
  - Think through the question step-by-step\\
  - Consider different ways the question could be understood\\
  - Look for potential ambiguities in terminology, scope, or intent\\
  - Consider how the provided context might influence interpretation\\\vspace{2.75mm}
  Task-Specific Instructions:\\
  \texttt{\textcolor{prompt_colour}{\{instructions\}}}\\\vspace{2.75mm}
  Output Format:\\
  Provide your analysis and reasoning, then conclude with interpretations and answers. Follow the format below:\\\vspace{2.75mm}

  ...analysis and reasoning...\\
  \textless answer\textgreater\\
  **Interpretation 1:** [State first interpretation]\\{} 
  [Answer for interpretation 1]\\\vspace{2.75mm}
  [Continue for all interpretations (up to 5) if the question is ambiguous...]\\
  \textless answer\textgreater
    \end{tcolorbox}

Task-specific instructions for question answering and text-to-SQL parsing are shown below: 
\begin{tcolorbox}[colback=white, colframe=prompt_colour2, coltitle=white,
                  title=\textbf{Answer Generation Instructions},
                  breakable, halign=left]
   \fontsize{8}{9.5}\selectfont
  - Provide direct, concise answers based on the given context. Keep your answers short and to the point.\\
  - Do not include any other text or explanation in the lines that contain the answers.
\end{tcolorbox}

\begin{tcolorbox}[colback=white, colframe=prompt_colour2, coltitle=white,
                  title=\textbf{SQL Generation Instructions},
                   halign=left]
  \fontsize{8}{9.5}\selectfont
  - Make sure you only output the information that is asked in the question. If the question asks for a specific column, make sure to only include that column in the SELECT clause, nothing more.\\
  - The generated query(ies) should return all of the information asked in the question without any missing or extra information.\\
  - Use a fenced code block with "sql" annotation, e.g. \verb|```sql ... ```|
\end{tcolorbox}

\section{DAPO Algorithm}
\label{appx:grpo_dapo}
DAPO \citep{yu2025dapo} extends GRPO with several key modifications designed for long-form reasoning tasks. It removes the KL penalty term (as divergence from the initial model is acceptable during reasoning training) and introduces the following changes:
\begin{asparaitem}
    \item[\bf Decoupled Clipping] DAPO uses asymmetric clipping parameters $\epsilon_{\text{low}}$ and $\epsilon_{\text{high}}$ to enhance exploration while maintaining stability.

\item[\bf Token-Level Loss] Instead of sample-level averaging, DAPO weights tokens equally across all responses, giving longer sequences more influence proportional to their length.

\item[\bf Dynamic Sampling] DAPO filters out prompts where all sampled responses are either correct or incorrect (accuracy equal to 1 or 0), ensuring effective gradients throughout training.

\item[\bf Overlong Reward Shaping] DAPO applies a soft length penalty that gradually reduces rewards for responses exceeding a maximum length threshold $L_{\max}$, rather than hard truncation. This discourages excessively long outputs while preserving useful gradient signal.
\end{asparaitem}

Let $G$ denote the number of responses sampled per prompt. Each response $o_i$ receives a binary correctness reward:
\begin{equation*}
    R_i = \text{is\_equivalent}(a, o_i) \in \{0, 1\}
\end{equation*}
which is then shaped via the overlong reward penalty:
\begin{equation*}
    R_i^{\text{shaped}} = R_i \cdot \max\!\left(1 - \alpha \cdot \frac{\max(|o_i| - L_{\max},\, 0)}{L_{\max}},\; \beta_{\min}\right)
\end{equation*}
where $L_{\max}$ is the maximum desired response length, $\alpha$ controls the penalty slope, and $\beta_{\min}$ is a floor preventing the reward from being fully zeroed out.

The DAPO objective is defined over a filtered subset $D' \subseteq D$ that excludes prompts with trivial outcomes:
\begin{equation*}
    D' = \left\{ (q, a) \in D \;\middle|\; 0 < \sum_{i=1}^{G} \mathbf{1}[\text{is\_equivalent}(a, o_i)] < G \right\}
\end{equation*}

The training objective is then:
\begin{equation*}
\begin{split}
    J_{\text{DAPO}}(\theta) &= \mathbb{E}_{(q,a)\sim D',\; \{o_i\}_{i=1}^G \sim \pi_{\theta_{\text{old}}}(\cdot|q)} \\
    &\quad \Bigg[ \frac{1}{\sum_{i=1}^G |o_i|} \sum_{i=1}^G \sum_{t=1}^{|o_i|} \min\!\big( r_{i,t}(\theta)\,\hat{A}_{i,t},\; \\
    &\qquad \text{clip}\!\left(r_{i,t}(\theta),\; 1 - \epsilon_{\text{low}},\; 1 + \epsilon_{\text{high}}\right) \hat{A}_{i,t} \big) \Bigg]
\end{split}
\end{equation*}
where the probability ratio and normalized advantage are given by:
\begin{equation*}
    r_{i,t}(\theta) = \frac{\pi_\theta(o_{i,t} \mid q,\, o_{i,<t})}{\pi_{\theta_{\text{old}}}(o_{i,t} \mid q,\, o_{i,<t})}
\end{equation*}
\begin{equation*}
    \hat{A}_{i,t} = \frac{R_i^{\text{shaped}} - \operatorname{mean}\!\left(\{R_j^{\text{shaped}}\}_{j=1}^G\right)}{\operatorname{std}\!\left(\{R_j^{\text{shaped}}\}_{j=1}^G\right)}
\end{equation*}

\section{Types of Ambiguity}
\label{appx:types_of_ambiguity}

Each dataset we use has its own ambiguity taxonomy. Below, we briefly describe the types of ambiguity as defined by the dataset authors.  For more details, we refer readers to the original papers.

\textbf{Abg-CoQA}~\cite{guo2021AbgCoQA} defines four ambiguity types:

\textit{Coreference resolution} occurs when pronouns or referring expressions can map to multiple entities in the context.  For example, given a passage mentioning a husband and a son, the question ``What is his name?'' could refer to either.

\textit{Time-dependency} occurs when temporal context is underspecified, allowing questions to refer to different time periods. For example, in a story describing an accident, the question ``What was everyone doing?'' can refer to actions either before or after the accident. The example in Table~\ref{tab:types_of_response} illustrates a similar case where ``What other state ruled it?'' can refer to different historical periods.

\textit{Answer types} are ambiguous with respect to the intended level of granularity. For example, when asked ``What pupils' were reading?'', the answer could be either novels a specific book. %

\textit{Event ambiguity} occurs when a story contains multiple parallel events or storylines, and questions underspecify which event they are referring to. For example, when a story mentions both a police investigation and CNN's reporting on the same incident, a question about whether the suspect was identified by name could refer to either storyline (the police identify but CNN does not).

\textbf{Ambrosia}~\citep{ambrosia} defines three ambiguity types: 

\textit{Scope ambiguity} occurs when it is unclear which elements a quantifier (e.g., ``every'', ``each'') refers to. For example, the question ``What activities does each gym offer?'' has a  collective interpretation (a common class for all gyms in the database) and distributive one (each gym is considered separately, a list of all classes and gyms).

\textit{Attachment ambiguity} occurs when it is unclear where a modifier attaches in a sentence. For example, ``Show screenwriters and editors on a work-for-hire'' can mean either both roles are on work-for-hire contracts (high attachment to the verb) or only editors are (low attachment to ``editors''). Table~\ref{tab:types_of_response} shows another example where a budget constraint may attach to one or both genres.

\textit{Vagueness} occurs when context creates uncertainty about which database entities are being referenced. For example, a question about ``required experience'' of the candidate is vague when the database contains both minimum and preferred years columns.

\textbf{AmbiQT}~\citep{bhaskar-etal-2023-benchmarking} defines four ambiguity types: 

\textit{Column ambiguity} occurs when multiple columns satisfy the same semantic description.
For example, the question ``List the ids of all students'' is ambiguous when a database contains both columns named \texttt{admission\_number} and  \texttt{roll\_number}.

\textit{Table ambiguity}  occurs when multiple tables represent similar entities.  For example,~``How many singers do we have?'' could query either an \texttt{artist} or \texttt{performer} table.

\textit{Join ambiguity} occurs when it is unclear whether information should be retrieved from a single table or from joining multiple tables. For example, asking for a person's name and address is ambiguous when \texttt{address} is in the same table as \texttt{name} and \texttt{postal\_address} is in a separate \texttt{person\_details} table. 

\textit{Precomputed aggregates} arise when databases store pre-aggregated values alongside raw data. For example, ``Find the average weight for each pet type'' could either compute the aggregate from data or select from  pre-computed \texttt{avg\_weight}.

\section{LLM-as-a-Judge Setup}\label{appx:judge}
We use an LLM judge because it can recognize semantic equivalence
despite wording differences unlike rule-based metrics (exact
match, word-overlap F1).
For example, in
Table~\ref{tab:types_of_response} ``The early Polish state'' vs.\ ``Poland''
would score 0\% F1, despite having identical meaning. This is
crucial for evaluating zero-shot and RL-trained models, which often
generate answers in different styles than gold annotations. We adapt the prompt from \citet{sun-etal-2024-head}, who found 98\% agreement between human judgments and LLM-based metrics.

We chose Qwen3 4B Instruct \citep{yang2025qwen3technicalreport} as the judge model because it is fast and cost-effective for large-scale evaluation, and provides reliable judgments for our task. 
In preliminary experiments, we compared our judge with the larger
Qwen3~235B~A22B Instruct, across multiple evaluation sets and found~1.4\%
difference in judgment accuracy on average. In most cases, the smaller
model was more conservative, marking particularly non-obvious answer
reformulations as incorrect.

\section{Training Hyperparameters}
\label{appx:training_details}
We train our models using DAPO implemented with the TRL library \citep{vonwerra2022trl} with hyperparameters adapted from TRL defaults.
We use the \texttt{Qwen3-4B-Instruct-2507} model and %
 train for 5 epochs with AdamW  (fused implementation) with learning rate 5e-7, linear scheduler with 3\% warmup,  and gradient clipping at norm 1.0. The per-device batch size is 16. During training, we generate 16 completions per prompt with maximum completion length of 2,500 tokens. 

We use vLLM \citep{kwon2023efficient}, Flash Attention 2 \citep{dao2023flashattention2} and DeepSpeed \citep{deepspeed} for efficient generation during training. All experiments were conducted on 5 NVIDIA H100 80GB GPUs: 1 GPU dedicated to the vLLM server for generation and 4 GPUs for training. 
For evaluation, we use a single GPU, and choose the last checkpoint. We use default sampling parameters (recommended by the developers) for Qwen3 4B Instruct: temperature = 0.7, top-k = 20, top-p = 0.8, and maximum length is 30k tokens.
This limit applies uniformly across all methods for fair comparison, and is
generous enough to accommodate the long traces of the Thinking baselines
(Figure~\ref{fig:coverage_vs_length}). Our model, however, rarely generates outputs
longer than the 2,500-token completion limit used during training.

Code, data, and models are available at
\url{https://github.com/saparina/intentRL}.

\section{Comparison Details}
\label{appx:comparison_details}

We use the updated versions of the Qwen3 model family: \texttt{Qwen3-4B-Instruct-2507}, \texttt{Qwen3-4B-Thinking-2507}, \texttt{Qwen3-235B-A22B-Instruct-2507}, \texttt{Qwen3-235B-A22B-Thinking-2507}.

We train SFT baselines using the TRL library with default hyperparameters. We modified the prompts shown in Appendix~\ref{appx:prompts} to exclude chain-of-thought instructions (e.g., ``Provide your analysis and reasoning'') since we do not have gold chain-of-thought reasoning.
Following the same procedure as our method, we train for 5 epochs, balance the training data, and select the last checkpoint for evaluation. We use the same model (Qwen3 4B Instruct) with the same LoRA configuration. Training uses the  AdamW optimizer (fused implementation) with a learning rate 2e-5, linear scheduler with no warmup, and gradient clipping at norm 1.0. The per-device batch size is 8 with no gradient accumulation. We use gradient checkpointing for memory efficiency and set the maximum prompt length to 8192 tokens. All SFT experiments were conducted on 1 NVIDIA H100 80GB GPU.

For both prompting and fine-tuned methods, we use the same maximum length as our method: 30k tokens. We also use the default sampling parameters recommended by the model developers for all models.

\section{Additional Results}\label{appx:additional_results}

\begin{table}[t]
\centering
\small
\begin{tabular}{llccccc}
\toprule
 & & \multicolumn{3}{c}{\textbf{Ambiguous}} & \multicolumn{2}{c}{\textbf{Unambiguous}} \\ 
\cmidrule(lr){3-5} \cmidrule(lr){6-7}
\textbf{Model} & \textbf{Method} & \textbf{Recall} & \textbf{Precision} & \textbf{Full Coverage} & \textbf{Recall} & \textbf{Precision}  \\ 
\midrule
\multirow{3}{*}{{Qwen3 4B Instruct}}   & IntentPrompt  & 22.5 & 30.8 & 3.8 & 22.0 & 18.8   \\
  & SFT         & 68.5 & 75.6 & 58.8 & 77.7 & \textbf{74.7} \\
  & IntentRL    & \textbf{82.4} & \textbf{77.5} & \textbf{74.1} & \textbf{88.7} & \textbf{74.7} \\ 
\midrule
\multirow{3}{*}{{Gemma3 4B IT}}  & IntentPrompt  & 21.8 & 26.0 &  2.3 & 42.1 &  25.6 \\
  & SFT         & 56.3 &  61.2 & 50.4 & 65.0 & 63.4 \\
  & IntentRL    & \textbf{78.8} & \textbf{72.8} & \textbf{69.5}  & \textbf{80.9} & \textbf{71.9}  \\
\midrule
\multirow{3}{*}{{Llama3.1 8B Instruct}}& IntentPrompt & 17.4 &  16.4 &  2.3 & 35.6 &   18.1 \\
  & SFT         & 70.5  &  70.9 & 64.9 & 75.7 &  74.2 \\
& IntentRL    & \textbf{79.4} & \textbf{74.4} & \textbf{71.8}  & \textbf{87.4} & \textbf{78.8}  \\
\bottomrule
\end{tabular}
\caption{Model family comparison on Ambrosia (Recall, Precision, and Full Coverage). Best results per model are in \textbf{bold}.}
\label{tab:model_families_ambrosia}
\end{table}
\paragraph{Generalization to Different Model Families}\label{appx:other_models}
To evaluate whether our findings generalize beyond Qwen models, we experiment with instruction-tuned Gemma3 4B and Llama3.1 8B (Table~\ref{tab:model_families_ambrosia}). Across models, IntentRL outperforms both IntentPrompt and SFT baselines. Gains are consistent across  model architectures and sizes, with improvements of 8--12\% in recall and \mbox{5--9\% }in full coverage over SFT. Prompting alone 
struggles across all models (13.9--22.5\% recall), demonstrating that explicit training for ambiguity resolution is necessary regardless of base model choice. Qwen3 4B Instruct achieves the strongest performance overall, consistent with its reported advantages \citep{yang2025qwen3technicalreport}, though all models benefit from our training approach. 
Performance differences across models suggest that stronger base models yield better results, while consistent improvements demonstrate that our approach is model-agnostic and represents a general methodology for handling ambiguity.

\begin{table}[t]
\centering
\begin{subtable}[t]{1\textwidth}
\centering
\small
\begin{tabular}{lccccccc}
\toprule
 &  \multicolumn{4}{c}{\textbf{Ambiguous}} & \multicolumn{3}{c}{\textbf{Unambiguous}} \\ 
\cmidrule(lr){2-5} \cmidrule(lr){6-8}
\cmidrule(lr){3-5}
 & \textbf{Overlap} & \textbf{Judge} & \textbf{Judge} & \textbf{Full} & \textbf{Overlap} & \textbf{Judge} & \textbf{Judge} \\ 
\textbf{Method}  & \textbf{F1} & \textbf{Recall} & \textbf{Precision} & \textbf{Coverage} & \textbf{F1} & \textbf{Recall} & \textbf{Precision} \\ 
\midrule
\multirow{1}{*}{IntentRL} 
 & 72.9 & 78.1 & 53.5 & 61.0  &  84.4 & 90.9	 &  40.7 \\
  \multirow{1}{*}{w/o Balancing} & 38.1 & 45.8 & 81.3 & 15.5 & 77.9 & 82.1 & 	82.1  \\
  \multirow{1}{*}{Down-sampling} & 62.3 & 75.4 & 43.7 & 59.4 & 86.2 & 90.8 & 37.4  \\
\bottomrule
\end{tabular}
 \caption{ Ablation of IntentRL on Abg-CoQA. }
\label{tab:Abg-CoQA_additional_results}
\end{subtable}
\hfill
\begin{subtable}[t]{1\textwidth}
\centering
\small
\bigskip
\begin{tabular}{lccccc}
\toprule
 & \multicolumn{3}{c}{\textbf{Ambiguous}} & \multicolumn{2}{c}{\textbf{Unambiguous}} \\ 
\cmidrule(lr){2-4} \cmidrule(lr){5-6}
 \textbf{Method} & \textbf{Recall} & \textbf{Precision} & \textbf{Full Coverage} & \textbf{Recall} & \textbf{Precision} \\ 
\midrule
\multirow{1}{*}{IntentRL} 
 & \textbf{82.4} & 77.5 & 74.1 & 88.7 & 74.7 \\
  \multirow{1}{*}{w/o Balancing}   & 49.4 & 85.0 & 9.2 & 84.5 & 81.7 \\
\bottomrule
\end{tabular}
\caption{
Ablation of IntentRL on Ambrosia.
}
\label{tab:ambrosia_additional_results}
\end{subtable}
\caption{Ablation of IntentRL: \textit{w/o Balancing} refers to IntentRL trained without training data balancing and \textit{Down-sampling} refers to balancing by removing unambiguous examples rather than oversampling ambiguous ones (see Section~\ref{sec:training-details})}.
\end{table}

\paragraph{Data Balancing Ablations}
We ablate the effect of training data balancing (Section~\ref{sec:training-details}). 
As shown in Tables~\ref{tab:Abg-CoQA_additional_results} and~\ref{tab:ambrosia_additional_results}, removing the 3:1 ambiguous-to-unambiguous ratio causes full coverage drops from 61.0\% to 15.5\% on Abg-CoQA and from 74.1\% to 9.2\% on Ambrosia, while precision increases from 53.5\%--77.5\% to over 80\% on both benchmarks. 
Without balancing,  the training set is dominated by unambiguous examples, which are easier, and the model already performs well on them by producing a single answer. DAPO reinforces this behavior since the reward signal from these straightforward examples is strong and consistent, while ambiguous examples are rare and harder to learn from. 

We also compare against down-sampling unambiguous examples to the same ratio on Abg-CoQA. Up-sampling is better across all metrics on ambiguous questions, with a particularly large gap in word-overlap F1 (72.9 vs.\ 62.3), while performance on unambiguous questions is comparable. Balancing is thus an important
design choice in our setup rather than a minor hyperparameter. We view the
underlying issue as an instance of the difficulty bias in GRPO-style training
\citep{liu2025understanding}; oversampling is one remedy, but alternative sampling
schemes, curricula, or modifications to the RL objective may work as well. We leave
this to future work.

\paragraph{SFT with Silver Interpretations} On Abg-CoQA, SFT predicts
answers only, as the dataset lacks gold interpretations. To verify this does not
disadvantage the baseline, we generated silver interpretations by prompting Qwen3
4B Instruct to convert each gold clarification question into an explicit
interpretation, and trained SFT on the resulting pairs. Results are slightly below
answer-only SFT on ambiguous questions (39.9 vs.\ 40.1 word-overlap F1, 17.9 vs.\
20.3 full coverage), consistent with Ambrosia, where the answer-only SFT variant
also outperforms the variant trained with gold interpretations
(Table~\ref{tab:ambrosia_results_full}). Interpretations vary more in wording and
are therefore harder to learn, so adding them does not improve SFT.

\paragraph{General Knowledge Evaluation}
To verify that IntentRL does not degrade general capabilities, we evaluate on standard reasoning and knowledge benchmarks: GSM8K \citep{cobbe2021gsm8k}, MMLU \citep{hendryckstest2021}, MMLU-Pro \citep{wang2024mmlu}, and MMLU-Redux \citep{gema2024mmlu}. All evaluations use 5-shot chain-of-thought prompts following \citet{yang2025qwen3technicalreport}. Note that we use the updated Qwen3 model (\texttt{Qwen3-4B-Instruct-2507}) rather than the version evaluated by \citet{yang2025qwen3technicalreport}. As shown in Table~\ref{tab:general_tasks}, IntentRL models match the base Qwen3 4B Instruct across all tasks, confirming that training for ambiguity resolution does not come at the cost of general performance. This is an encouraging signal that IntentRL could be integrated into broader post-training regimes alongside other reasoning domains such as mathematics or code, likely with larger datasets.

\begin{table}[t]
\centering
\small
\begin{tabular}{lcccc}
\toprule
 & \textbf{GSM8K} & \textbf{MMLU} & \textbf{MMLU-Pro} & \textbf{MMLU-Redux} \\ 
\midrule
Qwen3 4B Instruct (reported)~\tablefootnote{\url{https://huggingface.co/Qwen/Qwen3-4B-Instruct-2507}} & --- & --- & 69.6 & 84.2 \\ 
Qwen3 4B Instruct (reproduced) & 92.2 & 80.0 & 70.7 & 84.8 \\
IntentRL (Abg-CoQA) & 92.8  & 80.8  & 71.0 & 84.9 \\
IntentRL (Ambrosia) & 92.4 & 80.2 & 70.4 & 85.0\\
\bottomrule
\end{tabular}
\caption{
Accuracy of Qwen3 4B Instruct and IntentRL variants trained on Abg-CoQA and Ambrosia on mathematical reasoning (GSM8K) and general knowledge tasks (MMLU, MMLU-Pro, MMLU-Redux).}
\label{tab:general_tasks}
\end{table}

\paragraph{Comparison with State-of-the-Art on Out-of-Domain Data}
We extend the comparison with DisambigParse \citep{saparina-lapata-2025-disambiguate}, the current state of the art for ambiguous text-to-SQL, to the out-of-domain setting. \citet{saparina-lapata-2025-disambiguate} highlight out-of-domain generalization as a key strength of their approach. Table~\ref{tab:disambig_comparison_ambiqt} reports results on AmbiQT for both methods when trained on Ambrosia. IntentRL outperforms DisambigParse across most metrics, with substantial gains in precision (58.2\% vs.\ 35.3\%) and full coverage (49.1\% vs.\ 30.0\%), demonstrating that the advantages of IntentRL hold even in the setting most favourable to the previous state of the art.

\begin{table}[t]
\small
\centering
  \begin{tabular}{lcccc}
  \toprule
  \small
  \textbf{Method} & \textbf{Recall} & \textbf{Precision} & \textbf{Single Coverage} & \textbf{Full Coverage} \\ 
  \midrule
  DisambigParse & 58.5 & 35.3 & 88.0 & 30.0 \\
  IntentRL  & 66.9 & 58.2 & 84.7 & 49.1 \\
  \bottomrule
  \end{tabular}
  \caption{Comparison between DisambigParse \citep{saparina-lapata-2025-disambiguate} and IntentRL on AmbiQT when trained on Ambrosia. 
  We report Recall, Precision, Single and Full Coverage.}
  \label{tab:disambig_comparison_ambiqt}
\end{table}

\paragraph{Performance of Closed-Source Frontier Models}\label{appx:frontier_models}

To contextualise our results against the strongest available systems, we evaluate two closed-source frontier models: GPT-5.4~\footnote{\url{https://openai.com/index/introducing-gpt-5-4/}} (\texttt{gpt-5.4-2026-03-05}) and Gemini~2.5~Pro~\footnote{\url{https://blog.google/innovation-and-ai/models-and-research/google-deepmind/gemini-model-thinking-updates-march-2025/}}. Both are proprietary systems whose exact architectures and parameter counts are undisclosed, though they are widely understood to be substantially larger than our 4B parameter model. We use the same IntentPrompt template (Appendix~\ref{appx:prompts}) for both models. For GPT-5.4, we evaluate three reasoning effort levels (low, medium, high) to assess whether additional compute at inference time improves ambiguity handling. For all models, we set the maximum generation length to 30k tokens, matching the budget used for IntentRL and all other methods in our evaluation. Due to the cost of running closed-source API models at scale, we focus this comparison on Ambrosia, which provides verifiable evaluation through SQL execution and represents the more challenging setting for ambiguity resolution.

Table~\ref{tab:frontier_comparison_ambrosia} reports results on the ambiguous and unambiguous subsets of Ambrosia. IntentRL considerably outperforms both closed-source models across all metrics, achieving 74.1\% full coverage compared to 16.0\% for GPT-5.4 at high effort and 14.5\% for Gemini~2.5~Pro. Recall follows a similar pattern: 82.4\% for IntentRL versus 50.3\% for GPT-5.4 and 49.5\% for Gemini~2.5~Pro. Increasing reasoning effort from low to high results in only modest improvements for GPT-5.4 (8.4\% to 16.0\% full coverage), indicating that additional inference-time computation alone does not address the underlying challenge. On unambiguous questions, IntentRL also maintains a clear advantage in both recall (88.7\% vs.\ 82.2--83.8\% for GPT-5.4) and precision (74.7\% vs.\ 53.8--58.8\%). These results reinforce our central finding: handling ambiguity effectively requires targeted training rather than model scale or inference-time compute, particularly for structured output tasks where generating multiple valid alternatives demands systematic reasoning about the input.

\begin{table}[t!]
\centering
\small
\begin{tabular}{lccccc}
\toprule
 & \multicolumn{3}{c}{\textbf{Ambiguous}} & \multicolumn{2}{c}{\textbf{Unambiguous}} \\ 
\cmidrule(lr){2-4} \cmidrule(lr){5-6}
\textbf{Method} & \textbf{Recall} & \textbf{Precision} & \textbf{Full Coverage} & \textbf{Recall} & \textbf{Precision} \\ 
\midrule
Gemini 2.5 Pro & 49.5 & 51.8 & 14.5 & 75.1 & 51.8 \\
GPT-5.4 (low) & 45.6 & 53.2 & 8.4 & 78.0 & 58.8 \\
GPT-5.4 (medium) & 49.8 & 49.1 & 14.5 & 83.8 & 57.5 \\
GPT-5.4 (high) & 50.3 & 45.7 & 16.0 & 82.2 & 53.8 \\
\midrule
IntentRL (4B) & \textbf{82.4} & \textbf{77.5} & \textbf{74.1} & \textbf{88.7} & \textbf{74.7} \\
\bottomrule
\end{tabular}
\caption{Comparison with closed-source  frontier models on Ambrosia. GPT-5.4 reasoning effort levels are shown in parentheses. Best results per metric are in \textbf{bold}.}
\label{tab:frontier_comparison_ambrosia}
\end{table}
\begin{table}[t]
\centering
\small
\begin{tabular}{lcc}
\toprule
\textbf{Method} & \textbf{Ambiguous F1} & \textbf{Unambiguous F1} \\
\midrule
IntentPrompt & 6.3 & 9.7 \\
SFT & 11.2 & 16.1 \\
IntentRL & \textbf{13.6} & \textbf{16.5} \\
\bottomrule
\end{tabular}
\caption{F1 on  ambiguous and unambiguous subsets of \textbf{AmbigQA}
(closed-book). All methods use Qwen3 4B Instruct.
Best results per metric are in
\textbf{bold}.}
\label{tab:ambigqa}
\end{table}

\paragraph{Generalization to Open-Domain Question Answering}\label{appx:ambigqa}
We additionally report exploratory results on AmbigQA \citep{min-etal-2020-ambigqa},
an open-domain benchmark with multiple valid answers per question. Unlike the
datasets in our main experiments, AmbigQA provides no grounding context, so we
evaluate in a closed-book setting where the model must rely on parametric
knowledge. We follow the same training setup as Section~\ref{sec:training-details}
and report F1, the standard AmbigQA metric. As shown in Table~\ref{tab:ambigqa},
IntentRL outperforms both IntentPrompt and SFT on ambiguous and unambiguous
subsets. However, absolute F1 is low across all methods. Since our reward operates
at the answer level, the training signal in a closed-book setup is confounded by
parametric knowledge: a model that surfaces every valid interpretation still scores
poorly if it lacks the underlying facts, which is orthogonal to the ambiguity
reasoning we target. AmbigQA also poses practical challenges for our setup, as
roughly 7\% of questions admit more than five interpretations (up to 41),
an  interpretation space too large for our method. 

\section{Alternative Approaches to Handling Ambiguity}\label{appx:alternative_approaches}

Handling ambiguous questions is a challenging task, in part because 
there is no single format or interface that is universally optimal 
for users, applications, and models. Below we consider three 
alternative approaches to ambiguity handling: generating only answers 
without explicit interpretations, sampling answers in parallel, and 
generating clarification questions. We compare these approaches with 
our proposed interpretation-answer format and show how it can be 
integrated into each: it naturally generalizes to the 
answer-only format (which can also be viewed as an ablation of our 
approach), and it helps produce better clarifications and answers, 
demonstrating its utility in multi-stage pipelines. Finally, we 
discuss the trade-offs between these formats.

\subsection{Answer-Only Generation}
\label{appx:answer_only}

We consider approaches that generate multiple plausible answers 
$\{a_1, \dots, a_m\}$ without explicit interpretations. For the 
conversational QA example in Table~\ref{tab:types_of_response}, 
such approaches produce ``Greater Moravia'', ``The early Polish 
state'' and ``The Holy Roman Empire'' without disambiguating which 
interpretation each answer reflects. This can be viewed as an 
ablation of our approach: we train IntentRL to reason about user 
intent and generate answers only, without producing interpretations.

We consider the same comparisons with prompting alone and SFT. 
Note that in this setting, training does not require gold 
interpretations, only answers; this is the variant used for 
Abg-CoQA in Table~\ref{tab:Abg-CoQA_results}. We use the same 
prompt as for interpretation-answer pairs with minimal changes 
(removing the request for interpretations). Next we provide the prompt:

\begin{tcolorbox}[
  colback=white,
  colframe=prompt_colour,
  colbacktitle=prompt_colour,
  coltitle=white,
  title=\textbf{IntentPrompt Template, Answer-only}, breakable, halign=flush left
]
        \fontsize{8}{9.5}\selectfont
      You receive a context and a question. First analyse the question to identify different possible interpretations, then provide the corresponding answer for each interpretation.\\\vspace{2.75mm}
  If the question is unambiguous, provide a single answer. If the question contains ambiguities, underspecifications, or could be understood in multiple ways, provide multiple plausible answers (maximum of 5).\\\vspace{2.75mm}
  Context:\\
  \texttt{\textcolor{prompt_colour}{\{context\}}}\\\vspace{2.75mm}
  Question:\\
  \texttt{\textcolor{prompt_colour}{\{question\}}}\\\vspace{2.75mm}
  General Instructions:\\
  - Think through the question step-by-step\\
  - Consider different ways the question could be understood\\
  - Look for potential ambiguities in terminology, scope, or intent\\
  - Consider how the provided context might influence interpretation\\\vspace{2.75mm}
  Task-Specific Instructions:\\
  \texttt{\textcolor{prompt_colour}{\{instructions\}}}\\\vspace{2.75mm}
  Output Format:\\
  Provide your analysis and reasoning, then conclude with answers. Follow the format below:\\\vspace{2.75mm}

  ...analysis and reasoning...\\
    \textless answer\textgreater\\
   {}[State first answer]\\\vspace{2.75mm}
   {}[Continue for all answers (up to 5) if the question is ambiguous...]\\
  \textless answer\textgreater
    \end{tcolorbox}

\vspace{1em}
\paragraph{Results} Tables~\ref{tab:Abg-CoQA_results_full} and~\ref{tab:ambrosia_results_full} 
show results for answer-only generation alongside interpretation-answer 
pairs on Abg-CoQA and Ambrosia. Answer-only IntentRL is competitive 
with the interpretation-answer variant, showing comparable recall and full coverage on Abg-CoQA (78.9 vs.\ 78.1 and 61.8 vs.\ 61.0, respectively), 
though the interpretation-answer format achieves higher full coverage on Ambrosia  (74.1 vs.\ 71.8). This confirms that learning to reason about 
user intent using RL, not just the output format, is the primary driver 
of improvement. Both IntentRL variants outperform prompting and SFT in 
answer-only mode. The main advantage of the interpretation-answer format 
is interpretability: it provides explicit justifications for each answer, 
making predictions more transparent, while requiring the same supervision 
as the answer-only version. Providing interpretations is particularly 
beneficial in downstream tasks like text-to-SQL parsing, where the 
answer (an SQL query) can be difficult to understand without them.

\begin{table}[t]
\centering
\small
\begin{subtable}[t]{1\textwidth}
\begin{tabular}{@{}l@{\;\;\;}l@{\;\;\;}c@{\;\;}c@{\;\;}c@{\;\;}c@{\;\;\;}c@{\;\;}c@{\;\;}c@{}}
\toprule
 & & \multicolumn{4}{c}{\textbf{Ambiguous}} & \multicolumn{3}{c}{\textbf{Unambiguous}} \\ 
\cmidrule(lr){3-6} \cmidrule(lr){7-9}
\cmidrule(lr){4-6}
 & & \textbf{Overlap} & \textbf{Judge} & \textbf{Judge} & \textbf{Full} & \textbf{Overlap} & \textbf{Judge} & \textbf{Judge} \\ 
\textbf{Method} & \textbf{Model}  & \textbf{F1} & \textbf{Recall} & \textbf{Precision} & \textbf{Coverage} & \textbf{F1} & \textbf{Recall} & \textbf{Precision} \\ 
\midrule
\multicolumn{9}{@{}c}{\textit{Answer-only}} \\
\midrule
\multirow{4}{*}{{IntentPrompt}} 
 & 4B Instruct & 48.5 & 58.5  & 42.6 & 35.8 & 56.8 & 74.8 & 40.2  \\
&   4B Thinking & 31.3  & 60.4 & 52.6 & 37.4 & 53.4 & 72.7 & 50.2  \\
& 235B Instruct & 47.8 & 62.7 & 52.2 & 39.8 &  64.7 & 76.6 & 44.7 \\
 & 235B Thinking & 44.6 & 70.4 & 51.9 & 55.3 &  66.1 & 85.5 & 51.5 \\ 
\multirow{1}{*}{SFT} 
  & 4B Instruct & 40.1 &  48.1 & \textbf{70.3} & 20.3  & 72.2 & 77.6 &	\textbf{70.6} \\
   \multirow{1}{*}{IntentRL}  & 4B Instruct  & \textbf{73.8}  & \textbf{78.9} & 45.5 & \textbf{61.8} & \textbf{85.0} & 90.6 & 35.0 \\
\midrule
\multicolumn{9}{@{}c}{\textit{Interpretation-Answer Pairs}} \\
\midrule
\multirow{4}{*}{{IntentPrompt}} 
  & 4B Instruct & 44.4 & 62.3  & 42.7 &  45.5 & 59.2 & 77.7 & 40.6  \\
  & 4B Thinking & 18.2 & 46.2 & 45.3 & 31.7 & 32.5 & 60.6 & \textbf{44.6} \\
  & 235B Instruct & 55.2 & 63.1 & 47.0 & 45.5 & 68.9 & 80.8 & 42.1 \\
  & 235B Thinking & 47.8 & 66.2 & 49.7 & 50.4 & 59.6 & 82.1 & 47.2 \\
\multirow{1}{*}{IntentRL} 

  & 4B Instruct & \textbf{72.9} & \textbf{78.1} & \textbf{53.5} & \textbf{61.0}  &  \textbf{84.4} & \textbf{90.9}	 &  40.7 \\
\bottomrule
\end{tabular}
\caption{
Comparison of answer-only and interpretation-answer formats on Abg-CoQA.
}
\label{tab:Abg-CoQA_results_full}
\end{subtable}
\hfill
\begin{subtable}[t]{1\textwidth}
\bigskip
\begin{tabular}{llccccc}
\toprule
 & & \multicolumn{3}{c}{\textbf{Ambiguous}} & \multicolumn{2}{c}{\textbf{Unambiguous}} \\ 
\cmidrule(lr){3-5} \cmidrule(lr){6-7}
 \textbf{Method}& \textbf{Model} & \textbf{Recall} & \textbf{Precision} & \textbf{Full Coverage} & \textbf{Recall} & \textbf{Precision} \\ 
\midrule
\multicolumn{7}{@{}c}{\textit{Answer-only}} \\
\midrule
\multirow{4}{*}{{IntentPrompt}} 
& 4B Instruct & 23.8 & 48.1 & 2.3 & 58.3  & 57.2  \\
 &  4B Thinking &  27.4 & 49.1 & 2.3 & 58.3 &  56.1 \\
 &  235B Instruct & 35.2 & 53.3 & 8.4 & 69.3 &  64.1 \\
 & 235B Thinking & 28.2 & 32.8 & 5.3 & 61.8 &  48.4 \\ 
\multirow{1}{*}{SFT} 
  & 4B Instruct & 72.7 & 75.1 & 66.4 & 78.0 & 73.4 \\
\multirow{1}{*}{IntentRL}   & 4B Instruct  &  \textbf{82.4} &  \textbf{78.3} &  \textbf{71.8} & \textbf{86.1}  & \textbf{75.3} \\
\midrule
\multicolumn{7}{@{}c}{\textit{Interpretation-Answer Pairs}} \\
\midrule
\multirow{4}{*}{{IntentPrompt}} 
  & 4B Instruct & 22.5 & 30.8 & 3.8 & 22.0 & 18.8 \\
  & 4B Thinking & 16.9 & 23.2 & 3.1 & 25.2 & 19.7 \\
  & 235B Instruct & 46.2 & 48.9 & 18.3 & 74.1 & 57.8  \\
  & 235B Thinking & 37.3 & 37.8 & 12.2 & 53.7 & 35.9 \\
\multirow{1}{*}{SFT} 
  & 4B Instruct & 68.5 & 75.6 & 58.8 & 77.7 & \textbf{74.7} \\
\multirow{1}{*}{IntentRL} 
  & 4B Instruct & \textbf{82.4} & \textbf{77.5} & \textbf{74.1} & \textbf{88.7} & \textbf{74.7} \\
\bottomrule
\end{tabular}
\caption{
Comparison of answer-only and interpretation-answer formats on Ambrosia.
}
\label{tab:ambrosia_results_full}
\end{subtable}
\caption{
Comparison of answer-only and interpretation-answer formats on Abg-CoQA and Ambrosia. Best results per metric are in \textbf{bold} (marked separately per format).
}
\end{table}

\subsection{Parallel Sampling}
\label{appx:sampling}

Both our interpretation-answer pair generation and the answer-only 
generation considered above follow the same general approach: 
alternatives are generated sequentially within a single model call. 
The main advantage is that the model is conditioned on previously 
generated alternatives when producing the next one, which encourages 
it to explore distinct interpretations. 

An alternative strategy is to sample multiple independent responses 
from the model and aggregate them. If a question is ambiguous and 
the model recognises this, it should be less certain in its 
predictions, meaning alternatives should have comparable 
probabilities and thus appear across different samples. We refer 
to this approach as \textit{parallel sampling}. In this setting, 
the model is prompted with standard task-specific instructions 
(for conversational question answering and text-to-SQL parsing) 
with no mention of ambiguity or multiple outputs, which we refer 
to as the \textit{task prompt}. We provide these prompts below:

\begin{minipage}[t]{0.48\textwidth}
\begin{tcolorbox}[colback=white, colframe=prompt_colour2, coltitle=white,
                  title=\textbf{Conversational Question\\ Answering Task Prompt},
                   halign=left]
   \fontsize{8}{9.5}\selectfont
  You receive a context and a question. \\\vspace{2.75mm}
  Context:\\
  \texttt{\textcolor{prompt_colour}{\{context\}}}\\\vspace{2.75mm}
  Question to answer:\\
  \texttt{\textcolor{prompt_colour}{\{question\}}}\\\vspace{2.75mm}
  Answer Generation Instructions:
    \texttt{\textcolor{prompt_colour}{\{instructions\}}}\\\vspace{2.75mm}
  Output Format:\\
  Follow the format below:\\\vspace{2.75mm}
  \textless answer\textgreater\\{} 
  [State your answer]\\
  \textless answer\textgreater
\end{tcolorbox}
\end{minipage}
\hfill
\begin{minipage}[t]{0.48\textwidth}
\begin{tcolorbox}[colback=white, colframe=prompt_colour2, coltitle=white,
                  title=\textbf{Text-to-SQL Task Prompt},
                   halign=left]
  \fontsize{8}{9.5}\selectfont
  You receive a context and a question. \\\vspace{2.75mm}
  Context:\\
  SQLite database schema:\\
  \texttt{\textcolor{prompt_colour}{\{context\}}}\\
  This schema describes the database's structure, including tables, columns, primary keys, foreign keys, and any relevant relationships or constraints.\\\vspace{2.75mm}
  Question:\\
  \texttt{\textcolor{prompt_colour}{\{question\}}}\\\vspace{2.75mm}
  SQL Generation Instructions:
  \texttt{\textcolor{prompt_colour}{\{instructions\}}}\\\vspace{2.75mm}
  Output Format:\\
  Follow the format below:\\\vspace{2.75mm}
  \textless answer\textgreater\\{} 
  [State your SQL query]\\
  \textless answer\textgreater
\end{tcolorbox}
\end{minipage}

The model predicts a single answer per call, we then sample $n$ 
independent responses and aggregate all predicted answers. As the 
model is not expected to verbalise interpretations, we compare 
this approach to the answer-only variant of IntentPrompt. We also 
evaluate a variant without chain-of-thought reasoning (IntentPrompt 
w/o CoT), where the model is not prompted to provide explicit 
reasoning before generating answers. We consider two versions of 
parallel sampling: Sample-5 matches IntentPrompt's budget of 5 
answers, while Sample-10 doubles it. 

\paragraph{Results} Table~\ref{tab:Abg-CoQA_prompt_variations} and Table~\ref{tab:ambrosia_prompt_variations} provide results on Abg-CoQA 
and Ambrosia for different answer-only prompt variations.  All models 
are based on Qwen3 4B Instruct. With the task prompt, the model is 
not expected to predict more than a single answer; we also verified 
that it never responds with a clarification question (which could be 
a reasonable response to an ambiguous request) nor acknowledges 
ambiguity in any other way. Parallel sampling provides modest gains 
over a single response on Abg-CoQA and no gains on Ambrosia, and 
underperforms against IntentPrompt even with twice the computational budget 
(Sample-10). This indicates that the model consistently collapses to 
the same interpretation. In other words, \textbf{ambiguity is not inherently 
captured in the model's output distribution: our proposed approach 
learns to reason about intent and consider diverse interpretations rather than merely 
learning to format what parallel sampling could also recover}. 

Our findings are consistent with prior work showing that parallel sampling often 
fails to produce diverse outputs, as models tend to be biased towards 
certain completions, downweighting alternatives 
\citep{zhang2025noveltybench, troshin2025asking, zhang2025verbalized}. 
Moreover, parallel samples are not conditioned on each other, so 
there is no mechanism to encourage exploration of alternative 
interpretations. \citet{ambrosia} similarly considered $n$ completions 
from beam search as alternatives but found no advantage; our 
experiments extend their analysis with sampling (which is more diverse 
than beam search), and a more capable model (Qwen3 4B Instruct), yet reach the same conclusion. 

We note that our experiments use standard sampling parameters 
(temperature 0.7, top-k 20, top-p 0.8, as recommended by the model 
developers). More sophisticated test-time search strategies over the 
output space could potentially improve diversity, but this remains 
an open direction: to our knowledge, there are no established 
results demonstrating that such approaches successfully recover 
diverse interpretations of ambiguous inputs.

\begin{table}[t]
\centering
\begin{subtable}[t]{1\textwidth}
\centering
\small
\begin{tabular}{@{}l@{~}c@{~}c@{~~}c@{~~}c@{~~}c@{~~}c@{~~}c@{~~}c@{}}
\toprule
 &  & \multicolumn{4}{c}{\textbf{Ambiguous}} & \multicolumn{3}{c}{\textbf{Unambiguous}} \\ 
\cmidrule(lr){3-6} \cmidrule(lr){7-9}
\cmidrule(lr){4-6}
 & & \textbf{Overlap} & \textbf{Judge} & \textbf{Judge} & \textbf{Full} & \textbf{Overlap} & \textbf{Judge} & \textbf{Judge} \\ 
\textbf{Method} & \# answers & \textbf{F1} & \textbf{Recall} & \textbf{Precision} & \textbf{Coverage} & \textbf{F1} & \textbf{Recall} & \textbf{Precision} \\ 
\midrule
Task prompt & 1 &  30.3 & 35.0 & 62.6 & 10.6 & 51.0 & 66.5 & 67.6 \\
+Sample-5 & 5 & 40.6 & 41.9 & 62.0 & 17.9 & 55.5 & 72.4 & 67.7\\
+Sample-10 & 10 & 41.6 & 43.8 & 62.6 & 18.7 & 56.9 & 74.0 & 67.8 \\
IntentPrompt w/o CoT & 5& 29.3 & 41.5 & 43.0 & 14.6 & 48.8 & 70.2 & 52.0 \\
IntentPrompt  & 5& 48.5 & 58.5  & 42.6 & 35.8 & 56.8 & 74.8 & 40.2  \\ 
\bottomrule
\end{tabular}

\caption{Variations of \textit{answer-only} prompts on Abg-CoQA.}
\label{tab:Abg-CoQA_prompt_variations}
\end{subtable}
\hfill
\begin{subtable}[t]{1\textwidth}
\centering
\small
\bigskip
\begin{tabular}{@{}lcccccc@{}}
\toprule
 & & \multicolumn{3}{c}{\textbf{Ambiguous}} & \multicolumn{2}{c}{\textbf{Unambiguous}} \\ 
\cmidrule(lr){3-5} \cmidrule(lr){6-7}
 \textbf{Method} & \# answers  & \textbf{Recall} & \textbf{Precision} & \textbf{Full Coverage} & \textbf{Recall} & \textbf{Precision} \\ 
\midrule
Task prompt  & 1 &  21.2 & 49.6 & 0.0 & 56.3 & 56.3  \\
+Sample-5  & 5 &  24.2 & 50.5 & 0.0 & 65.4 & 60.2 \\
+Sample-10  & 10 & 24.2 & 50.5 & 0.0 & 66.7 &  60.2 \\
IntentPrompt w/o CoT & 5 & 16.0 & 36.5 & 0.8 & 42.4 & 42.4 \\
IntentPrompt  & 5 & 23.8 & 48.1 & 2.3 & 58.3  & 57.2  \\ 
\bottomrule
\end{tabular}
\caption{Variations of \textit{answer-only} prompts on Ambrosia.}
\label{tab:ambrosia_prompt_variations}
\end{subtable}
\caption{\label{tab:comparison}
Variations of \textit{answer-only} prompts. \textbf{Task Prompt} refers to prompting the model with standard task-specific instructions  with no mention of ambiguity. \textbf{+Sample-$n$} refers to sampling $n$ independent responses from the Task Prompt.
\textbf{IntentPrompt w/o CoT} refers to the prompt where we do not prompt the model to provide explicit reasoning. All models are based on Qwen3 4B Instruct.}
\end{table}

\subsection{Clarification Question Generation}
\label{appx:clarification}
Clarification question generation is a well-studied alternative 
approach for handling ambiguity: the model identifies ambiguous 
requests, generates a clarification question to help disambiguate, 
and returns it to the user for a response. Given the user's 
clarification (or after multiple rounds of interaction), the model 
then responds to the original request.

Our approach can be incorporated into such interactive pipelines in 
multiple ways. One natural option is to present interpretations (and 
possible answers) to the user and ask them to select one. This pairs 
well with clarification questions of the form ``Which of the following did you mean?'' and can be viewed as a variant of the interactive approach where the user is presented with a multiple-choice selection rather than providing an open-ended response.
While open-ended clarification in theory offers flexibility, it introduces practical challenges. For example, in Table~\ref{tab:types_of_response}, the user would need to respond ``No, I mean what state ruled it in the 14th century'' to obtain the Holy Roman Empire answer but the user may not know enough to answer precisely (e.g., when the user does not have access to the source passage). Alternatively, the user might simply respond ``no'', which narrows the interpretation space without fully disambiguating, potentially requiring multiple rounds. These challenges make clarification question generation an interesting but difficult problem to model and evaluate, especially without access to real users.

\begin{figure}
    \centering
    \includegraphics[width=\linewidth]{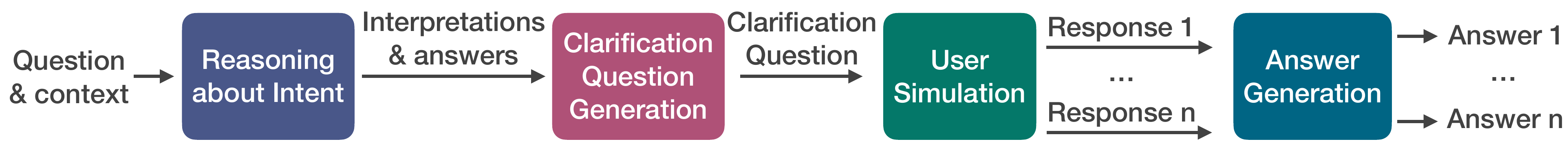}
    \caption{Open-ended clarification pipeline: (1) the model reasons about intent and provides possible interpretations and answers; (2) the model generates an open-ended clarification question; (3) the user simulator provides all possible responses to the clarification question; (4) for each response, the model generates a final answer. }
    \label{fig:clarification_pipiline}
\end{figure}

Following previous work, we consider user simulation with 
LLMs~\citep{Chen2024LearningTC}. We implement a multi-stage open-ended  clarification pipeline (Figure~\ref{fig:clarification_pipiline}):
\begin{enumerate}
    \item \textbf{Reasoning about Intent.} Given the question and context, the model reasons about possible interpretations and provides them along with candidate answers. This stage uses the approach proposed in this work. We consider three options: using IntentRL, using the original Qwen3 4B Instruct, or skipping this step entirely.

    \item \textbf{Clarification question generation.} The model generates an open-ended question whose answer would disambiguate the original ambiguous question. When interpretations and answers are available from the first stage, the model can rely on them; however, generating an effective clarification question remains non-trivial even with correct interpretations (e.g., \citet{kobalczyk2025active} define informative questions through information gain). We provide the model with one in-context example, although none of the models (including IntentRL) are specifically trained for clarification question generation. We consider both IntentRL and the original Qwen3 4B Instruct for this stage.

    \item \textbf{User simulation.}  We prompt a model to generate all possible user responses to the clarification question, simulating different users. This allows comparison with our approach. The quality of the user simulator is critical: if it misses plausible options or provides responses that fail to disambiguate, downstream performance suffers. We therefore use the largest model in the Qwen family, Qwen3 235B A22B Instruct and provide one in-context example.

    \item \textbf{Final answer generation.} For each simulated user response, we generate a final answer (again with one in-context example) using the same model as in the clarification stage (either IntentRL or the original Qwen3 4B Instruct). We then aggregate all answers across simulated responses and compute the metrics described in Section~\ref{sec:metrics}.
\end{enumerate}

We acknowledge that this pipeline is simplified: we do not train models for any particular stage (e.g., clarification question generation), and we do not include unambiguous questions, as that would require the model to first identify whether clarification is needed. We therefore experiment only with the ambiguous subsets. We provide prompts below.

\begin{tcolorbox}[
  colback=white,
  colframe=clarif_colour,
  colbacktitle=clarif_colour,
  coltitle=white,
  title=\textbf{Clarification Question Generation Instructions}, breakable, halign=flush left
]
\fontsize{8}{9.5}\selectfont
      You receive a context, a question, and possible interpretations of the question and their corresponding answers. You need to provide a clarification question that disambiguates the question.\\\vspace{2.75mm}
  Example:\\
  \texttt{\textcolor{clarif_colour}{\{example\}}}\\\vspace{2.75mm}
  Context:\\
  \texttt{\textcolor{clarif_colour}{\{context\}}}\\\vspace{2.75mm}
  Question:\\
  \texttt{\textcolor{clarif_colour}{\{question\}}}\\\vspace{2.75mm}
  Possible interpretations and answers:\\
  \texttt{\textcolor{clarif_colour}{\{interpretations\_and\_answers\}}}\\\vspace{2.75mm}
  General Instructions:\\
  - Ask a clarification question that will help to disambiguate the question.\\
  - Try to identify the single axis of ambiguity (scope, aggregation, filtering condition, entity reference, etc.).\\
  - Ask a clarification question about that axis so the user's reply naturally selects one interpretation.\\
  - Try to ask a concise clarification question that will be easy to understand and answer.\\\vspace{2.75mm}
  Output Format:\\
  Provide clarification question. Follow the format below:\\\vspace{2.75mm}
  \textless answer\textgreater\\{}
  [Clarification question]\\
  \textless /answer\textgreater
    \end{tcolorbox}

\begin{minipage}[t]{0.48\textwidth}
\begin{tcolorbox}[
  colback=white,
  colframe=user_sim_colour,
  colbacktitle=user_sim_colour,
  coltitle=white,
  title=\textbf{User Simulation Instructions}, halign=flush left
]
    \fontsize{8}{9.5}\selectfont
      You receive a context, a question, possible interpretations of the question and their corresponding answers and a clarification question. Provide all possible answers to the clarification question.\\\vspace{2.75mm}
  Example:\\
  \texttt{\textcolor{user_sim_colour}{\{example\}}}\\\vspace{2.75mm}
  Context:\\
  \texttt{\textcolor{user_sim_colour}{\{context\}}}\\\vspace{2.75mm}
  Ambiguous question:\\
  \texttt{\textcolor{user_sim_colour}{\{question\}}}\\\vspace{2.75mm}
  Possible interpretations and answers:\\
  \texttt{\textcolor{user_sim_colour}{\{interpretations\_and\_answers\}}}\\\vspace{2.75mm}
  Clarification question:\\
  \texttt{\textcolor{user_sim_colour}{\{clarification\_question\}}}\\\vspace{2.75mm}
  General Instructions:\\
  - Focus on covering all possible answers to the clarification question. They should cover all interpretations of the question.\\
  - The answers should be concise and to the point. They should be easy to understand and provide a clear disambiguation of the question.\\\vspace{2.75mm}
  Output Format:\\
  Provide all possible answers to the clarification question. Follow the format below:\\\vspace{2.75mm}
  \textless answer\textgreater\\
  Answer 1: [Answer 1 to the clarification question]\\
  Answer 2: [Answer 2 to the clarification question]\\
  ...\\
  Answer N: [Answer N to the clarification question]\\
  \textless /answer\textgreater

    \end{tcolorbox}
\end{minipage}
\hfill
\begin{minipage}[t]{0.48\textwidth}
\begin{tcolorbox}[
  colback=white,
  colframe=answer_gen_colour,
  colbacktitle=answer_gen_colour,
  coltitle=white,
  title=\textbf{Answer Generation Instructions},  halign=flush left
]
    \fontsize{8}{9.5}\selectfont
      You receive a context, a question, possible interpretations of the question and their corresponding answers, a clarification question, and the user's answer to that clarification question. Use the user's clarification answer to narrow down the interpretations and provide the final answer.\\\vspace{2.75mm}
  Example:\\
  \texttt{\textcolor{answer_gen_colour}{\{example\}}}\\\vspace{2.75mm}
  Context:\\
  \texttt{\textcolor{answer_gen_colour}{\{context\}}}\\\vspace{2.75mm}
  Ambiguous question:\\
  \texttt{\textcolor{answer_gen_colour}{\{question\}}}\\\vspace{2.75mm}
  Possible interpretations and answers:\\
  \texttt{\textcolor{answer_gen_colour}{\{interpretations\_and\_answers\}}}\\\vspace{2.75mm}
  Clarification question:\\
  \texttt{\textcolor{answer_gen_colour}{\{clarification\_question\}}}\\\vspace{2.75mm}
  User's answer to clarification:\\
  \texttt{\textcolor{answer_gen_colour}{\{clarification\_answer\}}}\\\vspace{2.75mm}
  General Instructions:\\
  - If the user's clarification answer matches one of the possible interpretations, provide the corresponding pair of interpretation and answer.\\
  - If none of the possible interpretations match the user's clarification answer, write a new interpretation and answer that matches the user's clarification.\\\vspace{2.75mm}
  Output Format:\\
  Provide the final interpretation and answer that matches the user's clarification. Follow the format below:\\\vspace{2.75mm}
  \textless answer\textgreater\\
  Interpretation: [State interpretation]\\{}
  [Answer]\\
  \textless /answer\textgreater
    \end{tcolorbox}
\end{minipage}

\paragraph{Results} Table~\ref{tab:comparison_pipiline} compares our Interpretation-Answer Pairs approach against the open-ended Clarification Pipeline across both datasets.

The top section of each subtable shows the direct Interpretation-Answer Pairs setting, which can also be viewed as the first stage of the pipeline in isolation. In the Clarification Pipeline results, we observe several patterns. First, generating clarification questions and answers without any prior reasoning about intent (first row) performs poorly, confirming that identifying possible interpretations is a valuable step to effective clarification. Second, incorporating interpretations from the base Qwen3 4B Instruct model helps on Abg-CoQA, where the initial interpretation quality is reasonable, but hurts on Ambrosia, where the base model's interpretations are too noisy to be useful. Third, using high-quality interpretations from IntentRL results in clear gains in both datasets. Using IntentRL for clarification and answer generation as well provides a further improvement, as IntentRL has also learned to predict answers more accurately; however, the dominant factor is the quality of the interpretations rather than the answer generation model.

\begin{table}[t]
\centering
\begin{subtable}[t]{1\textwidth}
\centering
\small
\begin{tabular}{llccccc}
\toprule
 \textbf{Reasoning} & \textbf{Clarification \& } & \textbf{Overlap} & \textbf{Judge} & \textbf{Judge} & \textbf{Full}  \\ 
 \textbf{about Intent} & \textbf{Answer Generation} & \textbf{F1} & \textbf{Recall} & \textbf{Precision} & \textbf{Coverage} \\ 
\midrule
\multicolumn{6}{@{}c}{\textit{Interpretation-Answer Pairs}} \\
\midrule
Qwen3 4B Instruct & --- & 44.4 & 62.3  & 42.7 &  45.5 \\
IntentRL & --- & \textbf{72.9} & \textbf{78.1} & \textbf{53.5} & \textbf{61.0} \\
\midrule
\multicolumn{6}{@{}c}{\textit{Open-ended Clarification Pipeline}} \\
\midrule
--- & Qwen3 4B Instruct & 42.6 & 57.3 & 38.7  & 36.6 \\
Qwen3 4B Instruct & Qwen3 4B Instruct & 50.6 & 61.2 & 46.5 & 41.5 \\
IntentRL & Qwen3 4B Instruct & 66.5 & 72.3 & 55.6 & 52.9  \\
IntentRL & IntentRL & 66.9 & 73.1  & 56.4 & 55.3  \\
\bottomrule
\end{tabular}
\caption{Comparison with clarification pipeline on Abg-CoQA (ambiguous questions).}
\label{tab:Abg-CoQA_pipiline}
\end{subtable}
\hfill
\begin{subtable}[t]{1\textwidth}
\centering
\small
\bigskip
\begin{tabular}{llcccc}
\toprule
  \textbf{Reasoning}  & \textbf{Clarification \& } &  & & \\ 
 \textbf{about Intent} & \textbf{Answer Generation}  & \textbf{Recall} & \textbf{Precision} & \textbf{Full Coverage} \\ 
\midrule
\multicolumn{5}{@{}c}{\textit{Interpretation-Answer Pairs}} \\
\midrule
Qwen3 4B Instruct & ---   & 22.5 & 30.8 & 3.8  \\
IntentRL & --- & \textbf{82.4} & \textbf{77.5} & \textbf{74.1} \\
\midrule
\multicolumn{5}{@{}c}{\textit{Open-ended Clarification Pipeline}} \\
\midrule
--- & Qwen 4B Instruct  & 36.3 & 38.8 & 8.4 \\
Qwen3 4B Instruct & Qwen3 4B Instruct  & 31.8 & 33.2 & 6.1 \\
IntentRL & Qwen3 4B Instruct & 74.6 & 71.7 & 59.5 \\
IntentRL & IntentRL & 76.2 & 75.0 & 63.4 \\
\bottomrule
\end{tabular}
\caption{Comparison with clarification pipeline on Ambrosia (ambiguous questions).}
\label{tab:ambrosia_pipiline}
\end{subtable}
\caption{\label{tab:comparison_pipiline} 
We compare our Interpretation-Answer Pairs approach, which eliminates the need for open-ended clarification, against a traditional Clarification Pipeline. For both settings, we ablate the choice of model (Qwen3 4B Instruct and IntentRL) across pipeline stages. For the Clarification Pipeline, we additionally ablate the inclusion of the reasoning stage (which provides interpretations).}
\end{table}

Notably, even the best Clarification Pipeline configuration underperforms the direct Interpretation-Answer Pairs approach with IntentRL. We manually examined typical failure cases and identified three main sources of error propagation: (1)~the clarification question targets the wrong axis of ambiguity or fails to disambiguate effectively, (2)~the user simulator does not cover all plausible responses, and (3)~both user simulation and answer generation can collapse to the same interpretation, missing valid alternatives.

These results suggest that open-ended clarification remains a hard problem that does not outperform our approach while requiring substantially more computation. Nonetheless, high-quality interpretations consistently improve clarification performance, indicating that our method is complementary to interactive pipelines. 
 Furthermore, our proposed approach could in principle be adapted to train clarification question generation directly: the model would reason about ambiguity but produce clarification questions instead of interpretation-answer pairs, with user simulation and answer generation incorporated into the reward computation.
We leave this direction for future work. 
More broadly, we believe the optimal strategy in practice would be a hybrid approach: presenting interpretations as multiple-choice options while allowing the user to provide an open-ended response if the listed interpretations do not cover their intended meaning.

\subsection{Discussion of Ambiguity Handling Strategies}\label{appx:format_tradeoffs}
The experiments presented in this section lead to three main conclusions about how our approach compares to alternative strategies for handling ambiguity.
 
\paragraph{Switching paradigms does not yield strong zero-shot results.}
A natural question is whether ambiguity handling can be achieved simply by adopting a different output format, that is, whether existing models already handle ambiguity well under the right prompting paradigm. Our results show this is not the case. Across all three alternative formats (answer-only generation, parallel sampling, and clarification question generation), zero-shot performance with prompting alone remains significantly below that of IntentRL.
These findings confirm that IntentRL does not merely train models to produce a particular output format when they already exhibit strong ambiguity awareness. Rather, the training itself is necessary: it teaches models to reason about user intent and systematically explore diverse interpretations.

\paragraph{No alternative format provides a clear advantage; each involves trade-offs.}
We find that none of the alternative paradigms offers a uniformly better solution than interpretation-answer pairs. Answer-only generation achieves competitive results, but sacrifices transparency: without interpretations, predictions lack justification, which is particularly problematic in downstream applications such as text-to-SQL where the generated query may be difficult to understand on its own. Parallel sampling avoids sequential conditioning but consistently collapses to the same interpretation, failing to recover the diversity that our approach achieves through explicit reasoning about alternatives. Open-ended clarification questions offer user control but introduce greater complexity: they require multiple interaction rounds, depend on the quality of user simulation, and suffer from error propagation across pipeline stages. Even the best clarification pipeline configuration underperforms direct interpretation-answer generation with IntentRL (Table~\ref{tab:comparison_pipiline}), with failures arising from poorly targeted clarification questions, incomplete user simulation, and answer collapse. A key limitation is the constrained number of interpretations (five in our work). For extremely vague questions with large interpretation spaces, clarification questions may be more appropriate. However, we believe the optimal strategy likely involves a hybrid: presenting interpretations as multiple-choice options while allowing users to provide open-ended responses when the listed options do not capture their intent.

\paragraph{Our approach integrates effectively with alternative formats.}
Despite the trade-offs of each paradigm in isolation, our approach integrates well with both the answer-only and clarification formats.
The interpretation-answer format naturally generalises to the answer-only setting, which can be viewed as an ablation where interpretations are simply omitted; both variants benefit from IntentRL training. More notably, high-quality interpretations from IntentRL consistently improve the clarification pipeline: incorporating IntentRL's reasoning about intent into the first stage of the pipeline leads to clear gains on both Abg-CoQA and Ambrosia, regardless of which model is used for subsequent clarification and answer generation (Table~\ref{tab:comparison_pipiline}). This demonstrates that learning to reason about user intent is complementary to interactive approaches rather than competing with them. In principle, our training framework could be extended to directly optimise clarification question generation, with user simulation and answer quality incorporated into the reward signal.
 
More broadly, an important open question is which response format is actually preferred by users, and whether the optimal format varies across tasks, domains, and interaction contexts. For instance, interpretation-answer pairs may be most useful in technical settings where transparency matters (e.g., text-to-SQL), while clarification questions may be preferred in conversational scenarios, and answer-only generation may suffice when brevity is prioritised. Our experiments evaluate these formats through automatic metrics, but user preferences may not align with metric-level performance. Conducting user studies to understand how people interact with different ambiguity handling formats, and whether the preferred format depends on factors such as expertise, task complexity, or time constraints, is an interesting direction for future work.

\section{Human Annotation Details}
\label{app:prolific}

For the interpretation-answer alignment evaluation, we recruited annotators through Prolific~\footnote{\url{https://www.prolific.com/}}  with the following criteria.
 We required annotators to have English as their primary language (and SQL knowledge for Ambrosia) and at least a 95\% approval rate on the platform. 
 Annotators were not allowed to retake the task.
To reduce cognitive load and make the task more efficient, we grouped interpretation-answer pairs by their source example. Each annotator evaluated 3 examples, with each example containing multiple interpretation-answer pairs from the same context.
We paid annotators £9 per hour, following Prolific's fair wage guidelines. The average completion time was 15 minutes.
We used the POTATO annotation tool \citep{pei-etal-2022-potato}.

\section{Limitations}\label{sec:discussion}

Our work has several limitations that suggest directions for future research.
 
\paragraph{Dataset coverage} While we demonstrate positive out-of-domain generalisation for text-to-SQL, we lack suitable out-of-domain conversational question answering datasets. Our approach requires multiple valid answers per input, which excludes datasets like PACIFIC~\citep{emnlp22-pacific}, where
each ambiguous turn is paired with a \emph{single} clarification question and its corresponding
answer in the following turn.
We also do not consider open-domain multi-answer datasets  such as AmbigQA~\citep{min-etal-2020-ambigqa} and ASQA~\citep{stelmakh-etal-2022-asqa}, because state-of-the-art systems for these datasets typically rely on retrieval (e.g., \citet{in-etal-2025-diversify}). This introduces orthogonal challenges beyond our focus on contextualised ambiguity resolution. Appendix~\ref{appx:ambigqa} reports exploratory closed-book results on AmbigQA, where IntentRL outperforms both baselines, though
absolute scores conflate method gains with  parametric knowledge.
Integrating IntentRL with retrieval-based approaches is a promising direction for future work.
 
\paragraph{Backbone model requirements} Like many RL-based methods, our approach requires a sufficiently strong backbone model. Weak models may fail to produce valid completions that receive non-zero rewards during training, preventing effective learning.  However, the model does not need to be good at ambiguity handling prior to training: on Ambrosia, the base model achieves only 3.8\% full coverage with prompting, while IntentRL improves this to 74.1\% , demonstrating that the ability to reason about ambiguity can be learned through training even when it is largely absent initially.

\paragraph{Inference cost} Our model produces all interpretations and
answers in \emph{one pass}, without a user in the loop, unlike clarification-based methods
which require an extra round of interaction to resolve the ambiguity, or two-stage pipelines that run a separate model
sequentially. This does not imply fewer tokens: non-reasoning methods can be more
token-efficient, and reasoning about intent may not be warranted for requests that
are clearly unambiguous.
 
\paragraph{Response format evaluation} We acknowledge that our interpretation-answer format is not universally optimal. It is constrained to a fixed number of interpretations (five in our work), which may be insufficient for highly vague questions with large interpretation spaces, particularly open-ended questions that lack the grounding context our
datasets provide. At the same time, it offers practical benefits: it provides explicit justifications for each answer, requires no additional annotation beyond multiple answers, and produces structured output that is easy to parse for downstream applications. We discuss further trade-offs between alternative formats in Appendix~\ref{appx:format_tradeoffs}.
We acknowledge that our comparison with alternative ambiguity handling formats is based entirely on automatic metrics. We do not conduct human studies to evaluate which format users actually prefer, or whether preferences vary across tasks, domains, and user expertise. As discussed in Appendix~\ref{appx:format_tradeoffs}, understanding user preferences for different response formats remains an open question and an important direction for future work.

\section{Model Outputs}
\label{appx:model_outputs}
Tables~\ref{tab:model_reasoning_trace_abgcoqa} and~\ref{tab:model_reasoning_trace_ambrosia} show full examples corresponding to those presented in Table~\ref{tab:types_of_response}, including model outputs with reasoning traces and interpretation-answer pairs.
In the conversational QA example (Table~\ref{tab:model_reasoning_trace_abgcoqa}), the model identifies all states that ruled Silesia in chronological order and generates five interpretations covering different time periods. The reasoning trace is notably repetitive, revisiting the same list of states multiple times before arriving at the final answer, illustrating a pattern we observe frequently in Abg-CoQA outputs (see Section~\ref{sec:analysis}).
In the text-to-SQL example (Table~\ref{tab:model_reasoning_trace_ambrosia}), the model correctly identifies the attachment ambiguity in ``horror films and thriller films with a budget of 20 million'' and generates SQL queries with different operator precedence. However, it also illustrates the redundancy issue discussed in Section~\ref{sec:analysis}: interpretations 1, 3, 4, and~5 express the same logical intent with only syntactic SQL variations, while the core ambiguity (whether the budget constraint applies to both genres or only thrillers) is captured by the difference between queries with and without parentheses in the \texttt{WHERE} clause. In practice, such redundant interpretations can be filtered out by comparing execution results.

\begin{table}[t]
\centering
\begin{tabular}{p{0.95\textwidth}}
\toprule
\fontsize{8}{9.5}\selectfont
\begin{minipage}{\linewidth}\fontsize{8}{9.5}\selectfont
Silesia is a region of Central Europe located mostly in Poland, with small parts in the Czech Republic and Germany\ldots{} The first known states to hold power there were probably those of \textcolor{cornell_red}{Greater Moravia} at the end of the 9th century and \textcolor{cornell_red}{Bohemia} early in the 10th century. In the 10th century, Silesia was incorporated into the \textcolor{cornell_red}{early Polish state}, and after its division in the 12th century became a \textcolor{cornell_red}{Piast duchy}. In the 14th century, it became a constituent part of the Bohemian Crown Lands under the \textcolor{cornell_red}{Holy Roman Empire}, which passed to the \textcolor{cornell_red}{Austrian Habsburg Monarchy} in 1526.

--- What are the other two countries it is part of? \hspace{3mm} --- the Czech Republic and Germany. \\
--- When did Bohemia rule there? \hspace{3mm} --- the 10th century
\end{minipage}
\textcolor{cornell_red}{\fontsize{8}{9.5}\selectfont --- What other state ruled it?} \\
\midrule
\multicolumn{1}{c}{\textit{\fontsize{8}{9.5}\selectfont Reasoning Trace}} \\
\midrule
{\fontsize{8}{9.5}\selectfont
\textbf{Analysis and reasoning:}}\vspace{2mm}

{\fontsize{8}{9.5}\selectfont The question ``What other state ruled it?'' is ambiguous because ``it'' refers to Silesia, which was ruled by multiple states over time. The previous question asked when Bohemia ruled there, and the context states that Bohemia ruled Silesia ``early in the 10th century.'' The question now asks for ``other states'' that ruled Silesia. The context mentions several states that ruled Silesia in sequence:}\vspace{1mm}

{\fontsize{8}{9.5}\selectfont
$\bullet$ Greater Moravia (end of 9th century)

$\bullet$ Bohemia (early 10th century)

$\bullet$ Early Polish state (10th century)

$\bullet$ Piast duchy (after division in 12th century)

$\bullet$ Bohemian Crown Lands under the Holy Roman Empire (14th century)

$\bullet$ Austrian Habsburg Monarchy (in 1526)}\vspace{2mm}

{\fontsize{8}{9.5}\selectfont The question ``What other state ruled it?'' implies that Bohemia is already known as a ruling state, and the user is asking for another state that ruled Silesia. Possible interpretations:}\vspace{2mm}

{\fontsize{8}{9.5}\selectfont 1. ``It'' refers to Silesia, and ``other state'' means other than Bohemia. The context states that Bohemia ruled Silesia ``early in the 10th century,'' and after that, Silesia was incorporated into the early Polish state. So, the early Polish state is a candidate. Also, later, it was ruled by the Austrian Habsburg Monarchy.}\vspace{2mm}

{\fontsize{8}{9.5}\selectfont 2. ``It'' could refer to a different entity. However, the previous context clearly refers to ``it'' as Silesia.}\vspace{2mm}

{\fontsize{8}{9.5}\selectfont 3. The phrasing ``what other state'' implies a plural or single additional state. The context states that Bohemia ruled Silesia early in the 10th century, and after that, it became a constituent part of the Bohemian Crown Lands under the Holy Roman Empire, which passed to the Austrian Habsburg Monarchy in 1526.}\vspace{2mm}

{\fontsize{8}{9.5}\selectfont Therefore, the states that ruled Silesia after Bohemia include: Early Polish state, Piast duchy, Bohemian Crown Lands under the Holy Roman Empire, Austrian Habsburg Monarchy.}\vspace{2mm}

{\fontsize{8}{9.5}\selectfont Key possible ``other states'' ruled Silesia: The early Polish state, The Piast duchy, The Holy Roman Empire (via Bohemian Crown Lands), The Austrian Habsburg Monarchy.}\vspace{2mm}

{\fontsize{8}{9.5}\selectfont Since the question is ambiguous, possible interpretations: ``What other state ruled it?'' could mean: What other state succeeded Bohemia? Answer: early Polish state. What other state ruled it after Bohemia? Answer: the early Polish state. What other state ruled it in the 14th century? Answer: the Holy Roman Empire. What other state ruled it after Bohemia? Answer: Austrian Habsburg Monarchy. What other state ruled it? (plausible answer: the early Polish state).}\vspace{2mm}

{\fontsize{8}{9.5}\selectfont From the context: ``In the 10th century, Silesia was incorporated into the early Polish state, and after its division in the 12th century became a Piast duchy. In the 14th century, it became a constituent part of the Bohemian Crown Lands under the Holy Roman Empire, which passed to the Austrian Habsburg Monarchy in 1526.''}\vspace{2mm}

{\fontsize{8}{9.5}\selectfont Thus, possible ``other states'' that ruled it include: early Polish state, Piast duchy, Holy Roman Empire (as ruler of Bohemian Crown Lands), Austrian Habsburg Monarchy.}\vspace{2mm}

{\fontsize{8}{9.5}\selectfont Possible interpretations of the question:
1. ``What other state ruled it?'' after Bohemia --- early Polish state.
2. What other state ruled it after Bohemia? --- the early Polish state.
3. What other state ruled it? (alternative to Bohemia) --- the early Polish state.
4. What other state ruled it? (after Bohemia) --- the Piast duchy.
5. What other state ruled it? (in the 14th century) --- the Holy Roman Empire.}\vspace{2mm}

{\fontsize{8}{9.5}\selectfont Given that the previous question was ``when did Bohemia rule there?'' and the answer was ``the 10th century,'' the current question is asking for another state that ruled Silesia. The most direct answer is the early Polish state, as it is the first state mentioned after Bohemia.}
\end{tabular}
\end{table}

\clearpage
\begin{table}[th!]
\centering
\begin{tabular}{p{0.95\textwidth}}
\fontsize{8}{9.5}\selectfont
{\fontsize{8}{9.5}\selectfont Other possible states: The early Polish state, The Piast duchy, The Holy Roman Empire, The Austrian Habsburg Monarchy.}\vspace{2mm}

{\fontsize{8}{9.5}\selectfont Final interpretations:
``What other state ruled it?'' --- refers to the state that ruled Silesia before Bohemia: Greater Moravia.
``What other state ruled it?'' --- refers to the state that ruled Silesia after Bohemia: early Polish state.
``What other state ruled it?'' --- refers to the state that ruled it in the 14th century: Holy Roman Empire.
``What other state ruled it?'' --- refers to the state that ruled it after Bohemian Crown Lands: Austrian Habsburg Monarchy.
``What other state ruled it?'' --- refers to the Piast duchy.}\vspace{2mm}

{\fontsize{8}{9.5}\selectfont The safest and most clear interpretation is that the question is ambiguous and ``it'' refers to Silesia, and the user is asking for another state that ruled it besides Bohemia. The context says Bohemia ruled early in the 10th century, and before that, Greater Moravia and Bohemia were early rulers. The state that ruled it after Bohemia was the early Polish state.}\vspace{2mm}

{\fontsize{8}{9.5}\selectfont Thus, possible answers: early Polish state, Piast duchy, Holy Roman Empire, Austrian Habsburg Monarchy, Greater Moravia.}\vspace{2mm}

{\fontsize{8}{9.5}\selectfont But the context says: ``The first known states to hold power there were probably those of Greater Moravia at the end of the 9th century and Bohemia early in the 10th century.'' So, Greater Moravia ruled before Bohemia.}\vspace{2mm}

{\fontsize{8}{9.5}\selectfont Therefore, the states that ruled Silesia in sequence:
1.~Greater Moravia,
2.~Bohemia,
3.~Early Polish state,
4.~Piast duchy,
5.~Bohemian Crown Lands under Holy Roman Empire,
6.~Austrian Habsburg Monarchy.}\vspace{2mm}

{\fontsize{8}{9.5}\selectfont ``Other state'' --- possible answers: Greater Moravia, early Polish state, Piast duchy, Holy Roman Empire, Austrian Habsburg Monarchy.}\vspace{2mm}

{\fontsize{8}{9.5}\selectfont Final interpretations: The question is ambiguous: ``it'' refers to Silesia, and ``what other state ruled it?'' could refer to any state besides Bohemia. Possible states: Greater Moravia, early Polish state, Piast duchy, Holy Roman Empire, Austrian Habsburg Monarchy.}\vspace{2mm}

{\fontsize{8}{9.5}\selectfont Since the context directly states: ``In the 10th century, Silesia was incorporated into the early Polish state, and after its division in the 12th century became a Piast duchy,'' these are clear.}\vspace{2mm}

{\fontsize{8}{9.5}\selectfont Thus, possible interpretations:
1. What other state ruled it (besides Bohemia) before the 10th century? --- Greater Moravia.
2. What other state ruled it after Bohemia? --- early Polish state.
3. What other state ruled it in the 12th century? --- Piast duchy.
4. What other state ruled it in the 14th century? --- Holy Roman Empire.
5. What other state ruled it after the Bohemian Crown Lands? --- Austrian Habsburg Monarchy.}\vspace{2mm}

{\fontsize{8}{9.5}\selectfont Given that the question is ambiguous, and based on the context, the intended meaning is likely to find the state that ruled Silesia directly after Bohemia, which is the early Polish state.}\vspace{2mm}

{\fontsize{8}{9.5}\selectfont However, the context states that Silesia was incorporated into the early Polish state in the 10th century. So, the answer ``early Polish state'' is valid.}\vspace{2mm}

{\fontsize{8}{9.5}\selectfont Thus, possible interpretations and answers:
``What other state ruled it?'' --- besides Bohemia --- Greater Moravia.
``What other state ruled it?'' --- besides Bohemia --- early Polish state.
``What other state ruled it?'' --- besides Bohemia --- Piast duchy.
``What other state ruled it?'' --- besides Bohemia --- Holy Roman Empire.
``What other state ruled it?'' --- besides Bohemia --- Austrian Habsburg Monarchy.}\vspace{2mm}

{\fontsize{8}{9.5}\selectfont Final selection (maximum 5 interpretations):}\\
\midrule
\multicolumn{1}{c}{\textit{\fontsize{8}{9.5}\selectfont Interpretation--Answer Pairs}} \\
\midrule
{\fontsize{8}{9.5}\selectfont \textbf{Interpretation 1:} The question refers to the state that ruled Silesia before Bohemia.} \vspace{-0.5mm}

{\fontsize{8}{9.5}\selectfont Greater Moravia}\vspace{2mm}

{\fontsize{8}{9.5}\selectfont \textbf{Interpretation 2:} The question refers to the state that ruled Silesia after Bohemia in the 10th century.}\vspace{-0.5mm}

{\fontsize{8}{9.5}\selectfont early Polish state}\vspace{2mm}

{\fontsize{8}{9.5}\selectfont \textbf{Interpretation 3:} The question refers to the state that ruled Silesia after becoming a Piast duchy.}\vspace{-0.5mm}

{\fontsize{8}{9.5}\selectfont Piast duchy}\vspace{2mm}

{\fontsize{8}{9.5}\selectfont \textbf{Interpretation 4:} The question refers to the Holy Roman Empire, which ruled Silesia in the 14th century.}\vspace{-0.5mm}

{\fontsize{8}{9.5}\selectfont Holy Roman Empire}\vspace{2mm}

{\fontsize{8}{9.5}\selectfont \textbf{Interpretation 5:} The question refers to the state that ruled Silesia after the Bohemian Crown Lands.}\vspace{-0.5mm}

{\fontsize{8}{9.5}\selectfont Austrian Habsburg Monarchy} \\
\bottomrule
\end{tabular}
\caption{Model output for an ambiguous conversational QA example from Abg-CoQA.}
\label{tab:model_reasoning_trace_abgcoqa}
\end{table}

\clearpage
\begin{table}[th!]
\centering
\begin{tabular}{p{0.95\textwidth}}
\toprule
\begin{minipage}{\linewidth}
\begin{lstlisting}[escapeinside=@@,aboveskip=0pt,belowskip=0pt,
basicstyle=\ttfamily\fontsize{6.7}{8}\selectfont\color{black},xleftmargin=0pt,framexleftmargin=0pt,breaklines=true]
CREATE TABLE Directors(
    id INTEGER PRIMARY KEY, name TEXT,
    birth_year INTEGER, nationality TEXT);
\end{lstlisting}
\vspace{1mm}
\begin{lstlisting}[escapeinside=@@,aboveskip=0pt,belowskip=0pt,
basicstyle=\ttfamily\fontsize{6.7}{8}\selectfont\color{black},xleftmargin=0pt,framexleftmargin=0pt,breaklines=true]
CREATE TABLE Film_Director(
    film_id INTEGER, director_id INTEGER,
    PRIMARY KEY(film_id, director_id));
\end{lstlisting}
\vspace{1mm}
\begin{lstlisting}[escapeinside=@@,aboveskip=0pt,belowskip=0pt,
basicstyle=\ttfamily\fontsize{6.7}{8}\selectfont\color{black},xleftmargin=0pt,framexleftmargin=0pt,breaklines=true]
CREATE TABLE Film_Genre(
    film_id INTEGER, @\textcolor{cornell_red}{genre\_id}@ INTEGER,
    PRIMARY KEY(film_id, genre_id));
\end{lstlisting}
\vspace{1mm}
\begin{lstlisting}[escapeinside=@@,aboveskip=0pt,belowskip=0pt,
basicstyle=\ttfamily\fontsize{6.7}{8}\selectfont\color{black},xleftmargin=0pt,framexleftmargin=0pt,breaklines=true]
CREATE TABLE Films(
    id INTEGER PRIMARY KEY, title TEXT,
    release_date DATE, director TEXT, @\textcolor{cornell_red}{budget}@ REAL);
\end{lstlisting}
\vspace{1mm}
\begin{lstlisting}[escapeinside=@@,aboveskip=0pt,belowskip=0pt,
basicstyle=\ttfamily\fontsize{6.7}{8}\selectfont\color{black},xleftmargin=0pt,framexleftmargin=0pt,breaklines=true]
CREATE TABLE Genres(
    id INTEGER PRIMARY KEY, @\textcolor{cornell_red}{genre}@ TEXT);
\end{lstlisting}
\end{minipage}
\textcolor{cornell_red}{\footnotesize Show horror films and thriller films with a budget of 20 million. Show them in one table.} \\
\midrule
\multicolumn{1}{c}{\textit{\footnotesize Reasoning Trace}} \\
\midrule
{\fontsize{8}{9.5}\selectfont
The question is:}\vspace{2mm}

{\fontsize{8}{9.5}\selectfont\textbf{``Show horror films and thriller films with a budget of 20 million. Show them in one table.''}}\vspace{2mm}

{\fontsize{8}{9.5}\selectfont Step-by-step analysis:}\vspace{2mm}

{\fontsize{8}{9.5}\selectfont\textbf{1. Identify the films to be shown:}
The question asks for ``horror films and thriller films''. In the context, ``Horror'' and ``Thriller'' are genres stored in the \texttt{Genres} table. Films are linked to genres via the \texttt{Film\_Genre} table, which connects \texttt{film\_id} to \texttt{genre\_id}. So, we must find films that belong to either the Horror (genre\_id = 1) or Thriller (genre\_id = 2) genre.}\vspace{2mm}

{\fontsize{8}{9.5}\selectfont\textbf{2. Filter by budget:}
``With a budget of 20 million'' --- this clearly refers to films where the \texttt{budget} = 20000000.0. The \texttt{Films} table has a \texttt{budget} column of type REAL. So, we are to filter films that have both: Genre = Horror or Thriller, Budget = 20,000,000.0}\vspace{2mm}

{\fontsize{8}{9.5}\selectfont\textbf{3. What does ``Show them'' mean?}
The phrase ``Show horror films and thriller films'' could mean: List the titles of the films that are either horror or thriller and have a budget of 20 million. Show only the titles (as the films themselves). The next part: ``Show them in one table'' is ambiguous: It might mean: ``Return the results in a single result set'' (i.e., combine both horror and thriller films with the budget condition into one output). It might mean: ``Show the films (titles) in a single table'' --- which could imply a query that returns titles from the Films table, filtered appropriately. Alternatively, it could be interpreted as: ``Bring together (join) the horror and thriller films that have the budget into a single table (i.e., a query that returns one list of titles)''. Another possible interpretation: ``Create a view or a result set that contains horror and thriller films with a budget of 20 million'' --- which again leads to filtering films by both genre and budget.}\vspace{2mm}

{\fontsize{8}{9.5}\selectfont\textbf{4. What exactly is being requested to be shown?}
The phrase ``Show horror films and thriller films'' could mean: Show titles of films that are horror or thriller --- i.e., union both. Or, show films that are both horror and thriller (intersection), but that seems less likely given ``and'' in the context of genres (as a list of films that belong to either). However, ``horror films and thriller films'' typically means films that are in either of those genres --- i.e., union. Also, the phrase ``with a budget of 20 million'' applies to both --- so we are filtering the horror and thriller films \emph{that also have a budget of 20 million}.}\vspace{2mm}

{\fontsize{8}{9.5}\selectfont\textbf{5. ``Show them in one table'':}
This could mean: Return the film titles in a single result set (a table with one column, the title). It may imply that the result should be treated as a single table (i.e., a simple SELECT query returning titles). Alternatively, it could be asking to join or combine the horror and thriller films with budget 20M into a single output --- which is simply returned in one query.}\vspace{2mm}

{\fontsize{8}{9.5}\selectfont\textbf{6. Scope of the data:}
The only data we can extract from the schema is the titles of films that: Are in genre Horror (id = 1) or Thriller (id = 2), Have a budget of 20,000,000.0. We must retrieve only the film titles, as ``show horror films'' likely means showing the titles of those films.}\vspace{2mm}

{\fontsize{8}{9.5}\selectfont\textbf{7. Ambiguity in ``films'' --- should it be all films or only those that are both horror and thriller?}
``Horror films and thriller films'' can be read as: Films that are horror films \textbf{and} thriller films (intersection) $\rightarrow$ films that belong to both genres. Films that are horror films \textbf{or} thriller films (union) $\rightarrow$ films in either genre. In everyday language, ``horror films and thriller films'' often means the collection of films in either category --- i.e., union. However, in database queries, ``and'' can sometimes denote intersection. So we must consider both possibilities.}

\end{tabular}
\end{table}
\clearpage
\begin{table}[th!]
\centering
\begin{tabular}{p{0.95\textwidth}}
{\fontsize{8}{9.5}\selectfont\textbf{8. Ambiguity in ``show them in one table'':}
Could mean: return the titles in a single column (a one-column table). Could mean: create a view or a combined listing (still just one query). It might simply mean: return the results as a single result set (which is standard in SQL).}\\\vspace{2mm}
\\\midrule
\multicolumn{1}{c}{\textit{\fontsize{8}{9.5}\selectfont Interpretation--Answer Pairs}} \\
\midrule
{\fontsize{8}{9.5}\selectfont \textbf{Interpretation 1:} Show the titles of horror films and thriller films that have a budget of 20 million, displaying them in a single table (i.e., a list of film titles).} 

{\fontsize{6.7}{8}\selectfont\texttt{SELECT f.title FROM Films f JOIN Film\_Genre fg ON f.id = fg.film\_id WHERE fg.genre\_id = 1 OR fg.genre\_id = 2 AND f.budget = 20000000.0;}}\vspace{2mm}

{\fontsize{8}{9.5}\selectfont \textbf{Interpretation 2:} Show the titles of films that are both horror and thriller (i.e., films in both genres) and have a budget of 20 million, in one table.}

{\fontsize{6.7}{8}\selectfont\texttt{SELECT f.title FROM Films f JOIN Film\_Genre fg ON f.id = fg.film\_id WHERE fg.genre\_id = 1 AND fg.genre\_id = 2 AND f.budget = 20000000.0;}}\vspace{2mm}

{\fontsize{8}{9.5}\selectfont \textbf{Interpretation 3:} Show the titles of horror and thriller films with a budget of 20 million, combined in one result set (i.e., union of horror and thriller films with that budget).}

{\fontsize{6.7}{8}\selectfont\texttt{SELECT f.title FROM Films f JOIN Film\_Genre fg ON f.id = fg.film\_id WHERE (fg.genre\_id = 1 OR fg.genre\_id = 2) AND f.budget = 20000000.0;}}\vspace{2mm}

{\fontsize{8}{9.5}\selectfont \textbf{Interpretation 4:} Show the titles of films that are either horror or thriller and have a budget of 20 million, in a single table (i.e., a combined list).}

{\fontsize{6.7}{8}\selectfont\texttt{SELECT f.title FROM Films f JOIN Film\_Genre fg ON f.id = fg.film\_id JOIN Genres g ON fg.genre\_id = g.id WHERE g.genre = 'Horror' OR g.genre = 'Thriller' AND f.budget = 20000000.0;}}\vspace{2mm}

{\fontsize{8}{9.5}\selectfont \textbf{Interpretation 5:} Show the titles of horror and thriller films with a budget of 20 million, in one table, using the film titles only.}

{\fontsize{6.7}{8}\selectfont\texttt{SELECT f.title FROM Films f JOIN Film\_Genre fg ON f.id = fg.film\_id WHERE fg.genre\_id IN (1, 2) AND f.budget = 20000000.0;}} \\
\bottomrule
\end{tabular}
\caption{Model output for an ambiguous text-to-SQL example from Ambrosia.}
\label{tab:model_reasoning_trace_ambrosia}
\end{table}

\clearpage 
\onecolumn

\end{document}